\documentclass[10pt]{article}

\usepackage{Sty/fmt}
\usepackage{Sty/mcr}
\usepackage{Sty/pkg}
\usepackage{Sty/thmE}
\usepackage{Sty/algE}
\usepackage{Sty/authblk}
\usepackage{Sty/colorset}

\graphicspath{ {./Fig/} }

\title{Conditional Selective Inference for Robust Regression and Outlier Detection using Piecewise-Linear Homotopy Continuation}

\date{\today}

\makeatletter
\def\@fnsymbol#1{\ensuremath{\ifcase#1\or
{*}\or 
{}\or 
{}\or 
\else\@ctrerr\fi}}
\makeatother

\author[1]{Toshiaki Tsukurimichi}
\author[1]{Yu Inatsu}
\author[1,2]{Vo Nguyen Le Duy}
\author[1,2]{Ichiro Takeuchi \thanks{Corresponding to: takeuchi.ichiro@nitech.ac.jp}}
\affil[1]{Nagoya Institute of Technology}
\affil[2]{RIKEN}

\begin{document}

\maketitle

\begin{abstract}
 \noindent
  In practical data analysis under noisy environment, it is common to first use robust methods to identify outliers, and then to conduct further analysis after removing the outliers. 
 In this paper, we consider the problem of statistical inference of the model estimated after outliers are removed from the data.
 This problem can be interpreted as post selection inference, and we study this problem in the context conditional selective inference (SI).
 In order to apply the conditional SI framework to inference after outlier removal, it is necessary to characterize the events of how the robust method identifies outliers.
 Unfortunately, the existing conditional SIs cannot be directly applied to our problem because they are applicable to the case where the selection events can be represented by linear or quadratic constraints.
 In this paper, we propose a conditional SI method for popular robust regressions such as Least-absolute-deviation (LAD) regression and Huber regression by introducing a new computational method using a convex optimization technique called homotopy method.
 We show that the proposed conditional SI method is applicable to a wide class of robust regression and outlier detection methods and has good empirical performance on both synthetic data and real data experiments.

\end{abstract}

\section{Introduction}
\label{sec:introduction}
A common practice in data analysis under the presence of outliers is that a \emph{robust} method is first used to detect outliers, and then the final model is fitted after the detected outliers are removed. 
Since the detected outliers themselves often reveal new insights, it is important to judge whether the detected outliers are truly deviated from the final model.
In this paper, we study regression problems under the presence of outliers, and propose a valid statistical inference method on the outliers detected by a class of commonly-used robust regression methods.
A fundamental difficulty in this problem is that outlier detection and model fitting are done on the same data.
Traditional statistical inference, which assumes that the statistical model and the target for which inferences are conducted must be fixed {\it a priori}, cannot be used for this problem. \par
Our basic approach is to interpret this problem as a \emph{selective inference (SI)} problem as suggested in \cite{chen2020valid}.
In the past decade, there have been several approaches suggested in the literature toward addressing SI problems~\cite{benjamini2005false,leeb2005model,leeb2006can,benjamini2009selective,potscher2010confidence,berk2013valid,lockhart2014significance,taylor2014post}.
A particularly notable approach that has received considerable attention in the past few years is \emph{conditional SI} introduced in the seminal paper by \cite{lee2016exact} for Lasso problem.
Since then, conditional SI has been extended to various directions~\cite{fithian2015selective,choi2017selecting,tian2018selective,chen2019valid,hyun2018post,loftus2014significance,loftus2015selective,panigrahi2016bayesian,tibshirani2016exact,yang2016selective,suzumura2017selective,terada2017selective,yamada2018post_a,yamada2018post_b,shimodaira2019selective,tanizaki2020computing,duy2020computing,duy2020quantifying}.
%
The basic idea of conditional SI is to conduct inferences conditional on the \emph{selection event}.
\cite{lee2016exact} showed that a valid inference conditional on the selection event can be constructed if the selection event is characterized by a set of linear inequalities.
Later, \cite{loftus2015selective} extended the framework to the cases where the selection event is characterized by a set of quadratic inequalities. 
\cite{chen2020valid} applied the approach in \cite{loftus2015selective} for conditional SI on the model fitted after outlier removal for a class of outlier detection methods in which the event is characterized by a set of quadratic inequalities.

Our contribution in this paper is to propose a conditional SI method for outliers when they are detected using a class of commonly-used robust regression methods~\cite{huber1973robust,bickel1973some,andrews1974robust,koenker1978regression,cleveland1979robust,zaman2001econometric,rousseeuw2005robust,maronna2019robust}.
In contrast to \cite{chen2020valid}, the outlier detection events by robust regression methods cannot be characterized by a set of linear/quadratic inequalities anymore. 
To circumvent the difficulty, we introduce a novel computational framework based on \emph{piecewise-linear homotopy continuation} method, which enables us to identify more complex outlier detection events.
As the main examples, we particularly focus on traditional but still commonly-used two robust regression methods: least-absolute-deviation (LAD) regression~\cite{andrews1974robust,bickel1973some,harvey1977comparison,hill1977two,koenker1978regression} and Huber regression~\cite{huber1973robust,huber2004robust}. 
Actually, the authors in \cite{chen2020valid} mentioned that Huber regression can be casted into Lasso problem (called \emph{Huberized Lasso}) by using the technique discussed in \cite{she2011outlier} and thus the conditional SI method for Lasso by \cite{lee2016exact} can be used for Huberized Lasso\footnote{Note that the authors in \cite{chen2020valid} point out the possibility of Huberized Lasso approach, but did not provide any numerical results. For comparison in \S\ref{sec:numerical_experiments}, we implemented it by ourselves.}.
However, as we clarify in \S\ref{subsec:discussion_on_related_works} and numerically demonstrate in \S\ref{subsec:exp_huberized_lasso}, the suggested approach comes with the loss of power due to over-conditioning of Lasso SI by \cite{lee2016exact}, while our approach is optimally conditioned and more powerful.

Our piecewise-linear homotopy continuation approach is inspired by several studies on \emph{regularization path}~\cite{osborne2000new,Efron04a,HasRosTibZhu04,RosZhu07,BacHecHor06,RosZhu07,Tsuda07,Lee07,Takeuchi09a,takeuchi2011target,Karasuyama11,hocking11a,Karasuyama12a,ogawa2013infinitesimal,takeuchi2013parametric}. 
For a class of model fitting procedures including LAD regression and Huber regression, it has been shown that the path of solutions is represented by piecewise-linear functions of the regularization parameter.
In this paper, we exploit the piecewise-linearity of the solution path along the direction of test-statistic for characterizing the outlier detection events of a class of commonly-used robust regression methods.
We are currently working on several other conditional SI problems using conceptually similar approach~\cite{duy2020parametric,duy2020quantifying,sugiyama2020more,duy2021more,das2021fast,das2021fast,duy2021exact,sugiyama2021valid}.

There have been many studies on outlier detection based on classical statistical theory in regression models \cite{srikantan1961testing,joshi1972some,ellenberg1973joint,ellenberg1976testing}. 
There have also been studies on testing for outliers based on statistical hypothesis testing. 
For example, \cite{srivastava1998outliers} and \cite{pan1995multiple} discuss the likelihood ratio test, which tests whether the individual of interest is an outlier or not, using the mean-shift model, which is a common model to describe the existence of outliers. 
Unfortunately, these classical inference methods for outliers are valid only when the target outliers are determined {\it a priori}.
If we apply these classical methods to the outliers detected by observing the data, they are \emph{anti-conservative}, i.e., false positive detection rate cannot be controlled at the desired significance level.
In order to control the false detection rate with these classical methods, multiple testing correction is indispensable.
For example, in Bonferroni correction, if the number of instances is $n$ and the number of detected outliers is $K$, the correction factor is ${n}\choose{K}$.
This correction factor can be extremely large unless the number of outliers $K$ is fairly small, which makes inferences too conservative. 

\section{Problem Statement}
\label{sec:problem_statement}
Consider a regression problem with the number of instances $n$ and the number of features $d$. 
We denote the observed dataset as $(X, \bm y)$ where $X \in \RR^{n \times d}$ is the design matrix and $\bm y \in \RR^n$ is the response vector. 
We assume that the observed response is a realization of the following random response vector 
\begin{align}
 \label{eq:data_model}
 \bm Y = (Y_1, \ldots, Y_n)^\top \sim N(\bm \mu, \Sigma), 
\end{align}
where ${\bm \mu} \in \RR^n$ is the unknown mean vector and $\Sigma \in \RR^{n \times n}$ is the covariance matrix which is known or estimable from independent data, and the design matrix $X$ is assumed to be non-random. 
Throughout the paper, the upper-case $\bm Y$ indicates random response vector, while the lower-case $\bm y$ indicates the observed response vector.

\subsection{Robust regression and outlier detection}
We are concerned with the case where a majority of the instances are well-approximated by a linear model, while there are several \emph{outliers} which are deviated from the linear model. 
A common practice in such a case is to use a robust regression estimator to detect outliers, and consider a linear model after removing the detected outliers. 
The goal of this paper is to make inference on the ``clean'' model after removing the outliers. 
We consider the following two-step approaches:
\begin{description}
 \item[step 1] fit a robust regression model and detect outliers; 
 \item[step 2] make inference on the linear model fitted without the detected outliers.
\end{description}
In the first step, a robust regression method is applied to the entire observed instances.
Let $\hat{\bm \beta}^R$ be the robust regression estimator by a certain robust regression method. 
Then, outliers are detected based on the \emph{robust residual}
\begin{align*}
 r^R_i = y_i - \bm x_i^\top \hat{\bm \beta}^R, i \in [n],
\end{align*}
where the notation $[n]$ indicates the set of natural numbers up to $n$. 
For example, if the magnitude of the robust residual is greater than a certain threshold, the instance is regarded as an outlier. 
In the second step, a linear model is fitted by least-square (LS) method by using the dataset from which the detected outliers are removed.
Let $\cO \subset [n]$ be the set of outliers detected in step 1, and define 
\begin{align*}
 X^{-\cO} := I_n^{-\cO} X, ~~~ \bm y^{-\cO} := I_n^{-\cO} \bm y
\end{align*}
where $I_n^{-\cO}$ is an $n$-by-$n$ diagonal matrix in which the $i$th diagonal entry is 0 if $i \in \cO$ and 1 otherwise. 
Then, the linear model coefficient vector obtained by LS fit with the clean dataset $(X^{-\cO}, \bm y^{-\cO})$ is 
\begin{align*}
 \hat{\bm \beta}^{-\cO} := (X^{-\cO})^+ \bm y^{-\cO},
\end{align*}
where the superscript $^+$ for a matrix indicates the pseudo-inverse.
 
\subsection{Statistical inference for detected outliers}
We are concerned with statistical inference on the model after removing the detected outliers. 
Specifically, we consider testing each of the detected outliers $i \in \cO$ is truly deviated from the model fitted after removing the set of outliers. 
We consider a statistical test for each of the detected outliers $i \in \cO$, but for notational simplicity, we omit the subscript $i$ if there is no ambiguity.
For each $i \in \cO$, we consider the following statistical test: 
\begin{align}
 \label{eq:statistical_test1}
 {\rm H}_0: \mu_i = \bm x_i^\top \bm \beta^{-\cO} ~{\rm vs.}~ {\rm H}_1: \mu_i \neq \bm x_i^\top \bm \beta^{-\cO}
\end{align}
where $\bm \beta^{-\cO}$ is the population LS estimate of the linear coefficient vector defined as
\begin{align}
 \label{eq:populationLS}
 \bm \beta^{-\cO} = (X^{-\cO})^+ I_{n}^{-\cO} \bm \mu. 
\end{align}
As a test-statistic for the above test, we employ
\begin{align}
 \label{eq:test-statistic}
 Z = Y_i - \bm x_i^\top \hat{\bm \beta}^{-\cO}.
\end{align}
Throughout the paper, we do not assume that the linear model is correctly specified, i.e., it is possible that $\bm \mu \neq X^{-\cO} \bm \beta$ for any $\bm \beta \in \RR^{p}$. 
Even when the linear model is not correctly specified, $\bm \beta^{-\cO}$ in \eq{eq:populationLS} is still a well-defined best linear approximation. 
This model is often called \emph{saturated model}~\cite{fithian2015selective} in the context of conditional SI.

If we additionally assume that $\bm \mu$ in \eq{eq:data_model} follows a location-shift model in the form of 
\begin{align}
\label{eq:location-shift-model}
 \bm \mu = X \bm \beta + \bm u, 
\end{align}
where $\bm \beta \in \RR^p$ and $\bm u \in \RR^n$ such that $u_i \neq 0$ if the $i$th instance is outlier and 0 otherwise, and further assume that the first step correctly identified all the outliers, then the statistical test in \eq{eq:statistical_test1} is equivalent to 
\begin{align}
 \label{eq:null-alternative-hypotheses}
 {\rm H}_{0}: u_i = 0 ~~~{\rm vs.}~~~ {\rm H_{1}}: u_i \neq 0,
\end{align}
i.e., testing whether the $i$th instance is an outlier or not under the location-shift model in \eq{eq:location-shift-model}.
 
\subsection{Conditional selective inference (SI)}
In the statistical test \eq{eq:statistical_test1}, because the target of the inference is selected based on the observed data $\bm y$, if we naively apply a traditional statistical inference as if the inference target is pre-determined, the result is not valid any more.
To address the issue, we consider conditional SI framework as in \cite{chen2020valid}.
Let us now write the set of instances detected as outliers in step 1 as $\cO(\bm Y)$ in order to clarify that the outliers are detected based on the response vector $\bm Y$. 
The goal of conditional SI is to quantify the sampling property of the test statistic \emph{conditional} on the selection event $\cO(\bm Y) = \cO(\bm y)$ which indicates the event that the detected outliers by applying the robust regression method to the random response vector $\bm Y$ is the same as those for the actual observed response $\bm y$.
Specifically, we consider so-called \emph{selective $p$-value} $p$ which has the following sampling property under the null hypothesis ${\rm H}_{0}$: 
\begin{align}
 \label{eq:selective_p_property}
 {\rm Pr}(p < \alpha \mid \cO(\bm Y) = \cO(\bm y)) = \alpha, \forall~\alpha \in (0, 1).
\end{align}
Similarly, we consider $(1-\alpha)$ \emph{selective confidence interval (CI)} which has $(1 - \alpha)$ coverage conditional on the selection event.

\subsection{Formulation of selective $p$-values}
In order to formulate the selective $p$-values having the sampling property \eq{eq:selective_p_property}, we first note that the test-statistic in \eq{eq:test-statistic} is written as 
\begin{align*}
 Z = \bm \eta^\top \bm Y, \text{ where } \bm \eta := \left(I_n - X \left(X^{-\cO}\right)^+ I_n^{-\cO} \right)^\top \bm e(i),
\end{align*}
where $\bm e(i)$ is the $i$th standard basis vector in $\RR^n$.
Then, let us decompose the random response vector into the component which is uncorrelated with, and hence independent of, the test statistic $\bm \eta^\top \bm Y$ and the remaining component as follows:
\begin{align*}
 \bm Y = \bm q(\bm Y) + \Sigma \bm \eta (\bm \eta^\top \Sigma \bm \eta)^{-1} \bm \eta^\top \bm Y, 
\end{align*}
where the first term in the right hand side 
\begin{align*}
 \bm q(\bm Y) := (I_n - \Sigma \bm \eta (\bm \eta^\top \Sigma \bm \eta)^{-1} \bm \eta^\top) \bm Y 
\end{align*}
is the component independent of the test statistic $\bm \eta^\top \bm Y$. 
Let us consider a subset of data space 
\begin{align}
 \label{eq:conditional_data_space_y}
 \cY := \left\{\bm y^\prime \in \RR^n \mid \cO(\bm y^\prime) = \cO(\bm y), \bm q(\bm y^\prime) = \bm q(\bm y) \right\},
\end{align}
where the condition $\bm q(\bm y^\prime) = \bm q(\bm y)$ indicates that the nuisance component for a response vector $\bm y^\prime$ is the same as that for the observed response vector $\bm y$.

Then, the sampling distribution of the test statistic $\bm \eta^\top \bm Y$ conditional on $\bm Y \in \cY$ is characterized as
\begin{align}
 \label{eq:truncated-normal-definition}
 \bm \eta^\top \bm Y \mid \bm Y \in \cY \sim {\rm TN}(\bm \eta^\top \bm \mu, \bm \eta^\top \Sigma \bm \eta, \cZ), 
\end{align}
where ${\rm TN}(m, s^2, \cZ)$ indicates the truncated Normal distribution with the mean $m$, the variance $\sigma^2$, and the truncation region $\cZ$ (The truncation region $\cZ$ will be defined later in \eq{eq:cZ})
By denoting the c.d.f. of the truncated Normal distribution as $F_{\bm \eta^\top \bm \mu, \bm \eta^\top \Sigma \bm \eta}^{\cZ}$, we can obtain the following pivotal quantity that has the property
\begin{align}
 \label{eq:truncated-normality}
 F_{\bm \eta^\top \bm \mu, \bm \eta^\top \Sigma \bm \eta}^{\cZ}(\bm \eta^\top \bm Y) \mid \bm Y \in \cY \sim {\rm Unif}[0, 1],
\end{align}
where ${\rm Unif}[0, 1]$ indicates the uniform distribution in $[0, 1]$. 
Since the pivotal quantity in \eq{eq:truncated-normality} has ${\rm Unif}[0, 1]$ distribution for any realization of the component $\bm q(\bm Y)$, the pivotal quantity also has a ${\rm Unif}[0, 1]$ distribution marginally over all possible $\bm q(\bm Y)$, i.e., 
\begin{align}
 \label{eq:truncated-normality2}
 F_{\bm \eta^\top \bm \mu, \bm \eta^\top \Sigma \bm \eta}^{\cZ}(\bm \eta^\top \bm Y) \mid \{\cO(\bm Y) = \cO(\bm y)\} \sim {\rm Unif}[0, 1],
\end{align}
Using the pivotal quantity, the selective $p$-value for the test \eq{eq:null-alternative-hypotheses} is defined as 
\begin{align}
 \label{eq:selective-p-values}
 p = 2 \min\{\pi, 1-\pi\}, \text{ where } \pi = 1 - F_{0, \bm \eta^\top \Sigma \bm \eta}^{\cZ}(\bm \eta^\top \bm Y).
\end{align}
From \eq{eq:truncated-normality2}, we can confirm that the selective $p$-value $p$ in \eq{eq:selective-p-values} satisfies the desired property in \eq{eq:selective_p_property}.
Furthermore, to obtain $(1-\alpha)$ confidence interval for any $\alpha \in [0, 1]$, by inverting the pivotal quantity in \eq{eq:truncated-normality2}, we can find the smallest and largest values of $\bm \eta^\top \bm \mu$ such that the value remains in the interval $\left[\alpha/2, 1 - \alpha/2\right]$.
For more information on the contents of this subsection, see \S5 in \cite{lee2016exact} in which the authors derive the truncation region $\cZ$ when the selection event is represented as a polyhedron. 

\subsection{Conditional data space in a line}
The second condition $\{\bm q(\bm y^\prime) = \bm q(\bm y)\}$ in \eq{eq:conditional_data_space_y} indicates that the response vector $\bm y^\prime \in \cY$ is restricted on a line in $\RR^n$.
Consider a set of response vectors written as 
\begin{align*}
 \bm y^\prime = \bm a + \bm b Z, Z \in \RR, 
\end{align*}
where $\bm a := \bm q(\bm y)$ and $\bm b := \Sigma \bm \eta (\bm \eta^\top \Sigma \bm \eta)^{-1}$.
Then, for any $z^\prime \in \RR$, 
\begin{align*}
 \bm q(\bm a + \bm b z^\prime)
 &
 =
 (I_n - \Sigma \bm \eta (\bm \eta^\top \Sigma \bm \eta)^{-1} \bm \eta^\top) (\bm a + \bm b z^\prime)
 \\
 &
 =
 (I_n - \Sigma \bm \eta (\bm \eta^\top \Sigma \bm \eta)^{-1} \bm \eta^\top) ((I_n - \Sigma \bm \eta (\bm \eta^\top \Sigma \bm \eta)^{-1} \bm \eta^\top) \bm y + \Sigma \bm \eta (\bm \eta^\top \Sigma \bm \eta)^{-1} z^\prime)
 \\
 &
 =
 (I_n - \Sigma \bm \eta (\bm \eta^\top \Sigma \bm \eta)^{-1} \bm \eta^\top) \bm y
 \\
 &
 =
 \bm q(\bm y). 
\end{align*}
Therefore, the subset of random response vectors $\cY$ in \eq{eq:conditional_data_space_y} is restricted on a line and parametrized by a scalar parameter $z^\prime \in \RR$ as 
\begin{align*}
 \cY
 =
 \{
 \bm a + \bm b z^\prime
 \}_{z^\prime \in \cZ}, 
\end{align*}
where
\begin{align}
 \label{eq:cZ}
 \cZ := \{z^\prime \in \RR \mid \cO(\bm a + \bm b z^\prime) = \cO(\bm y)\}.
\end{align}
Note that the truncation region  $\cZ$ is the truncation region of the truncated Normal distribution in \eq{eq:truncated-normal-definition}. 

The above discussion indicates that the conditional SI is reduced to the problem of identifying the truncation region $\cZ$ in \eq{eq:cZ}.
Most of the existing conditional SI studies consider the case where the selection event is represented as a polyhedron in the data space. 
In this case, the truncation region $\cZ$ in \eq{eq:cZ} is represented as a single line segment and can be identified based on \emph{Polyhedral Lemma} (see \S5 in \cite{lee2016exact}). 
However, the selection event we consider in this study cannot be represented as a single polyhedron, we cannot rely on the computational method in these existing studies.
For computing the truncation region $\cZ$ for more complicated selection events, we introduce an optimization technique called piecewise-linear homotopy computation in \S4.

\section{Robust Regression and Outlier Detection}
\label{sec:robust_regression_and_outlier_detection}
In this paper, we mainly consider two robust regression estimators and two outlier detection criteria for which the proposed PLH-based conditional SI can be applied. 
In \S~\ref{subsec:discussion_applicable}, we will discuss to which class of robust regression estimators and outlier detection criteria the proposed PLH-based conditional SI can be applied.

\subsection{Robust Regression Estimators}
We consider two traditional yet most commonly used robust regression estimators. We note that we are concerned only with the robustness against the outliers in the response $y_i$ but not in the design variables $\bm x_i$, $i \in [n]$.

\paragraph{Least absolute deviation (LAD) regression}
The least absolute deviation (LAD) regression~\cite{andrews1974robust,bickel1973some,harvey1977comparison,hill1977two,koenker1978regression} is formulated as
\begin{align}
 \label{eq:LAD1}
 \hat{\bm \beta}^R
 :=
 \argmin_{\bm \beta}
 \sum_{i=1}^n
 \psi(y_i - \bm x_i^\top \bm \beta) 
 \text{ with }
 \psi(r) = |r|.
\end{align}
In contrast to least-square (LS) regression, LAD regression is known to be robust against outliers in the response. 
For solving the optimization problem in \eq{eq:LAD1}, iterative approach such as iterative re-weighted LS method is required.
It has been known that the optimization problem in \eq{eq:LAD1} is written as a linear programming problem, and a variant of simplex method can be used for effectively solving LAD regression~\cite{barrodale1973improved}.

\paragraph{Huber regression}
The Huber regression is formulated as
\begin{align}
 \label{eq:Huber1}
 \hat{\bm \beta}^R
 :=
 \argmin_{\bm \beta}
 \sum_{i=1}^n
 \psi(y_i - \bm x_i^\top \bm \beta) 
 \text{ with }
 \psi(r)
 =
 \mycase{
 \frac{1}{2} r^2
 &
 \text{ if } |r| \le \delta,
 \\
 \delta (|r| - \frac{1}{2} \delta)
 &
 \text{ otherwise},
 }
\end{align}
where $\delta > 0$ is a predetermined tuning parameter.
Huber loss $\psi$ in \eq{eq:Huber1} is one of the most commonly-used robust loss function. 
For solving the optimization problem in \eq{eq:Huber1}, iterative approach such as iteratively re-weighted LS method is used.
Later, we will exploit the fact that the optimization problem in \eq{eq:Huber1} is written as a quadratic programming problem. 

\subsection{Outlier Detection Criteria}
\label{subsec:outlier_detection_criteria}
We consider two simple methods for outlier detection.
The two methods is simply based on the residual from the robust regression estimator.
Remember that we denote the robust estimator as $\hat{\bm \beta}^R \in \RR^p$ and the robust residual of the $i$th instances $r_i^R = y_i - \bm x_i^\top \hat{\bm \beta}^R, i \in [n]$.

\paragraph{Threshold-based outlier detection}
The first outlier detection method is based on a predefined threshold $\xi > 0$.
We simply define the instance to be outlier if the magnitude of the robust residual is greater than the threshold, i.e., the set of outliers is defined as 
\begin{align}
 \label{eq:threshold_based_outlier_detection}
 \cO := \left\{
 i \in [n] \mid |r_i^R| \ge \xi
 \right\}, 
\end{align}
where $\xi > 0$ is a predetermined threshold. 

\paragraph{Top-$K$ outlier detection}
The second outlier detection method is based on the rank of the magnitude of the robust residuals.
Let $K < n$ be a pre-defined number of outliers, which is possible when one has prior knowledge about the fraction of outliers.
An instance is defined as an outlier if the magnitude of the residual is within the top $K$ out of $n$ instances, i.e., 
\begin{align}
 \label{eq:topK_outlier_detection}
 \cO := \left\{ i \in [n] \mid \mathop{\rm rank}_{i \in [n]}(|r_i^R|) \ge K \right\},
\end{align}
where $\mathop{\rm rank}_{i \in [n]}(|r_i^R|)$ is the rank of the $i$th absolute residual in the descending order of $\{|r_i^R|\}_{i \in [n]}$.

\section{Proposed Method: Piecewise-Linear Homotopy (PLH)-based Conditional SI}
\label{sec:sec4}
As we described in \S\ref{sec:problem_statement}, the main task of our conditional SI is reduced to the problem of identifying the truncation region $\cZ$ in \eq{eq:cZ}. 
To this end, we consider robust regression estimators for all possible response vectors written written as $\bm y(z) = \bm a + \bm b z$ for all possible $z \in \RR$.
In this section, we thus also denote all the other relevant quantities as functions of $z \in \RR$ such as $\hat{\bm \beta}^R(z) := \argmin_{\bm \beta} \sum_{i=1}^n \psi(y_i(z) - \bm x_i^\top \bm \beta)$, $r^R_i(z):= y_i(z) - \bm x_i^\top \hat{\bm \beta}^R(z), i \in [n]$, and $\cO(z) := \cO(\bm y(z))$ for all possible $z \in \RR$.
Here, note that the notation $\bm y$ still indicate the actual observed response vector, while $\bm y(z)$ for some $z \in \RR$ can be interpreted as other realization of the random response vector $\bm Y$.

In this paper, we concentrate our attention on the class of robust regression estimators, i.e., the class of loss functions $\psi$,  which can be represented as \emph{piecewise-linear} functions of $z$. 
We will discuss that a wide class of robust regression estimators including the two examples of robust regression estimators described in \S\ref{sec:robust_regression_and_outlier_detection} actually have this property. 

Let $z_0 < z_1 < \ldots < z_{T-1} < z_T \in \RR$ be the sequence of \emph{breakpoints} where $T$ is the number of breakpoints and we define $z_0 := -\infty$ and $z_T := +\infty$ for notational simplicity.
We denote the piecewise-linear form of robust regression estimators as 
\begin{align}
 \label{eq:piecewise-linear-solution}
 \hat{\bm \beta}^{R}(z)
 =
 \mycase{
 \bm c_1 + \bm d_1 z & \text{ if } z \in (z_0, z_1], \\
 \bm c_2 + \bm d_2 z & \text{ if } z \in [z_1, z_2], \\
 ~~~~ \vdots & ~~~~~~~ \vdots \\
 \bm c_T + \bm d_T z & \text{ if } z \in [z_{T-1}, z_T)
 }
\end{align}
using $T$ pairs of $p$-dimensional vectors $\{(\bm c_t, \bm d_t)\}_{ t \in [T]}$ which are  obtained by solving a family of convex optimization problems for all possible $z \in \RR$. 
With this class of robust regression estimators, it is easy to note that the robust residual vector $\bm r^R(z)$ is also represented as a piecewise-linear function of $z$ which we denote as 
\begin{align}
 \label{eq:piecewiselinear-residual}
 \bm r^R(z) := \bm y(z) - X \bm \beta^R(z) =
 \mycase{
 \bm f_1 + \bm g_1 z & \text{ if } z \in (z_0, z_1], \\
 \bm f_2 + \bm g_2 z & \text{ if } z \in [z_1, z_2], \\
 ~~~~ \vdots & ~~~~~~~ \vdots \\
 \bm f_T + \bm g_T z & \text{ if } z \in [z_{T-1}, z_T),
 }
\end{align}
 where
 \begin{align*}
  \bm f_t := \bm a - X \bm c_t
  \text{ and }
  \bm g_t := \bm b - X \bm d_t, t \in [T].
 \end{align*}
Since the truncation region $\cZ$ in \eq{eq:cZ} is written as a union of of such a region in all the intervals as
\begin{align}
 \cZ = \bigcup_{t=1}^T \left\{z \in [z_{t-1}, z_t] \mid \cO(\bm y(z)) = \cO(\bm y) \right\}, 
\end{align}
we only need to be able to characterize outlier detection event when the residuals linearly change with $z$. 
 
\subsection{Piecewise-linear Homotopy}
In order to show that the path of robust regression estimators $\hat{\bm \beta}^R(z)$ for the entire $z \in \RR$ in the two commonly-used methods discussed in \S\ref{sec:robust_regression_and_outlier_detection} are represented as piecewise-linear functions of $z$, we resort on the results established in the literature of \emph{parametric programming}~\cite{Ritter84,Allgower93,Gal95,Best96}. 
Parametric programming is a type of optimization methods whose goal is to solve a family of optimization problems parametrized by one or multiple parameters. 
Especially, in the contexts of linear programming and quadratic programming, parametric programming has been extensively studied and used in various problems. 
When a part of the objective function or constraints of a linear or quadratic programming problem is parametrized by a scalar parameter, it is known that the optimal solution path is a piecewise linear function of the parameter.
Our idea is to show that the problem of finding the path of robust regression estimators for $z \in \RR$ can be formulated as a parametric programming problem with the parameter $z$. 
In what follows, we show that LAD regression and Huber regression are respectively formulated as a linear and a quadratic parametric programming problems with the parameter $z$.

Intuitively, one may think that the truncation region $\cZ$ can be approximated by considering a fine sequence of grid points in the line and check whether the selection event happens or not at each of these grid points.
Unfortunately, this intuitive approach rarely works because it is impossible to know in advance how many fine grid points should be considered to obtain a sufficiently good approximation.
In fact, some truncation regions identified by the parametric programming approach contain very short line segments, which indicates that a large number of very fine grid points must be considered for good approximation.
On the other hand, it is computationally expensive to conduct the robust regression and the outlier detection at each of the large number of grid points. 
Therefore, a trade-off between approximation accuracy and computational cost should be considered for the grid points-based approach.
Since the parametric programming approach presented in this study can circumvent these two difficulties, we should always use it wherever applicable.
For problems cannot be represented as parametric LP nor QP, a possible direction is to employ recently developed technique so-called approximate parametric programming~\cite{giesen2012approximating,giesen2012approximating2,NIPS2015_82b8a343,ndiaye2019safe,NEURIPS2019_9e3cfc48}.

\paragraph{Parametric linear programming formulation of LAD regression}
It is well known that LAD regression is formulated as a linear programming problem~\cite{barrodale1973improved}.
We show that the LAD regression problem in \eq{eq:LAD1} can be formulated as a parametric linear programming problem with the parameter $z$ which appears in the constant part of the constraints, which enables us to use the known results of parametric linear programming whose solution path is represented as a piecewise-linear function of $z$ (see, e.g., Section 8.6 in \cite{murty1983linear}).
\begin{lemm}
 \label{lemm:lad_regression}
 The LAD regression problem
 \begin{align}
  \label{eq:LAD}
  \hat{\bm \beta}^R(z) := \argmin_{\bm \beta} \sum_{i=1}^n | y_i(z) - \bm x_i^\top \bm \beta|
 \end{align}
 is formulated as the following linear parametric programming problem: 
 \begin{align}
  \nonumber
  \min_{\bm r}
  ~ 
  &
  \bm q^\top \bm r
  \\
  \label{eq:parametric_linear_program1}
  {\rm s.t.}
  \;\;
  &
  S \bm r = \bm u_0 + \bm u_1 z, \bm r \ge \bm 0.
 \end{align}
 where $\bm r$ is the vector of variables, whereas  $S$, $\bm q$, $\bm u_0$ and  $\bm u_1$ are constant matrix/vectors for defining the optimization problem (their concrete definitions are presented in Appendix A).
 The path of the optimal solutions of the parametric linear programming problems in \eq{eq:parametric_linear_program1} is represented as a piecewise-linear function of $z$(see, e.g., \cite{RosZhu07}). 
\end{lemm}
\noindent
The proof of Lemma~\ref{lemm:lad_regression} is presented in Appendix A.

\paragraph{Parametric quadratic programming formulation of Huber regression}
It is well known that Huber regression is formulated as a quadratic programming problem. 
We show that the Huber regression problem in \eq{eq:Huber1} can be formulated as a parametric quadratic programming problem with the parameter $z$ which appears in the constant part of the constraints.
This property enables us to use the known results of parametric quadratic programming whose solution path is represented as a piecewise-linear function of $z$.
\begin{lemm}
 \label{lemm:huber_regression}
 The Huber regression problem
 \begin{align*}
  \hat{\bm \beta}^R(z) := \argmin_{\bm \beta} \sum_{i=1}^n \psi(y_i(z) - \bm x_i^\top \bm \beta) \text{ where }
  \psi(r)
  =
  \mycase{
  \frac{1}{2} r^2
  &
  \text{ if } |r| \le \delta,
  \\
  \delta (|r| - \frac{1}{2} \delta)
  &
  \text{ otherwise},
  }
 \end{align*}
 is formulated as the following quadratic parametric programming problem: 
 \begin{align}
  \nonumber
  \min_{\bm r}
  ~ 
  &
  \frac{1}{2} \bm r^\top P \bm r + \bm q^\top \bm r
  \\
  {\rm s.t.}
  \;\;
  &
  S \bm r \le \bm u_0 + \bm u_1 z.
  \label{eq:parametric_quadratic_program}
 \end{align}
 where $\bm r$ is the vector of variables, whereas $P$, $S$, $\bm q$, $\bm u_0$ and $\bm u_1$ are constant matrices/vectors for defining the optimization problem.
 The path of the optimal solutions of the parametric quadratic programming problems in \eq{eq:parametric_quadratic_program} is represented as a piecewise-linear function of $z$. 
\end{lemm}
\noindent
The proof of Lemma~\ref{lemm:huber_regression} is presented in Appendix A.
Furthermore, see Appendix B for the computation of the breakpoints in Huber regression.

\subsection{Outlier detection criteria}
Since the robust residuals are represented as a piecewise linear function of $z$, outlier detection events based on the robust residuals can be characterized in a simple form. 
Here, we show that the selection event $\cO(\bm y(z)) = \cO(\bm y)$ in a line $\{\bm y(z) \in \RR^n \mid \bm y(z) = \bm a + \bm b z, z \in \RR\}$ for the two simple outlier detection methods discussed in \S\ref{sec:robust_regression_and_outlier_detection} can be represented in a form that can be easily computed by a simple algorithm. 
The basic idea is to characterize the outlier detection event in each interval of two consecutive breakpoints $[z_{t-1}, z_t], t \in [T]$ by exploiting the fact that the robust residuals are linear in $z$ within each interval. 

\paragraph{Threshold-based outlier detection}
For threshold-based outlier detection in \eq{eq:threshold_based_outlier_detection}, the truncation region $\cZ$ in \eq{eq:cZ} is characterized as the following lemma. 
\begin{lemm}
 \label{lemm:threshold_based_outlier_detection}
 For $t \in [T]$ and $i \in [n]$, let
 \begin{align*}
  \cV_{t, i}
  =
  \mycase{
  \left[
  z_{t-1}, \min\left\{z_{t}, \frac{- \xi - f_{t,i}}{g_{t,i}}\right\}
  \right]
  \cup
  \left[
  \max\left\{z_{t-1}, \frac{\xi - f_{t,i}}{g_{t,i}}\right\}, z_t
  \right]
  &
  \text{ if }
  g_{t,i} > 0,
  \\
  \left[
  z_{t-1}, \min\left\{z_{t}, \frac{\xi - f_{t,i}}{g_{t,i}}\right\}
  \right]
  \cup
  \left[
  \max\left\{z_{t-1}, \frac{-\xi - f_{t,i}}{g_{t,i}}\right\}, z_t
  \right]
  &
  \text{ if }
  g_{t,i} <  0,
  \\
  \left[
  z_{t-1}, z_{t}
  \right]
  &
  \text{ if }
  g_{t,i} = 0
  \text{ and }
  |f_{t,i}| \ge \xi,
  \\
  \emptyset
  &
  \text{ if }
  g_{t,i} = 0
  \text{ and }
  |f_{t,i}| < \xi ,
  }
 \end{align*}
 where, with a slight abuse of notation, we regard an interval $[\ell, u]$ is empty if  $\ell > u$.
 Then, the truncation region $\cZ$ in \eq{eq:cZ} is given as
 \begin{align*}
  \cZ =
  \bigcup_{t=1}^T
  \left(
  \left(
  \bigcap_{i \in \cO(\bm y)}
  \cV_{t, i}
  \right)
  \cap
  \left(
  \bigcap_{i \notin \cO(\bm y)}
  [z_{t-1}, z_t] \setminus \cV_{t, i}
  \right)
  \right)
 \end{align*}
\end{lemm}
\noindent
The proof of Lemma~\ref{lemm:threshold_based_outlier_detection} is given in Appendix A.

\paragraph{Top $K$ outlier detection}
For top $K$ outlier detection in \eq{eq:topK_outlier_detection}, the truncation region $\cZ$ in \eq{eq:cZ} is characterized as the following lemma. 
\begin{lemm}
 \label{lemm:topK_outlier_detection}
 For $t \in [T]$ and $(i, i^\prime) \in \cO(\bm y) \times \left([n] \setminus \cO(\bm y)\right)$, let
 {\footnotesize
\begin{align*}
  \cW_{t, (i, i^\prime)}
  =
  \mycase{
 \left[
  \max\left\{z_{t-1}, \frac{-\gamma}{\beta}\right\}, z_t
  \right]
  &
  \text{ if }
 \alpha=0 \text{ and } \beta>0,
  \\
  \left[
  z_{t-1}, \min\left\{z_{t},\frac{-\gamma}{\beta}\right\}
  \right]
  &
  \text{ if }
 \alpha=0 \text{ and } \beta<0,
 \\
 \left[z_{t-1}, z_{t}
 \right]
  &
  \text{ if }
 \alpha=0 \text{ and } \beta=0 \text{ and } \gamma\ge0,
\\
\emptyset
  &
  \text{ if }
  \alpha=0 \text{ and } \beta=0 \text{ and } \gamma<0,
\\
 \left[
  \max\left\{z_{t-1}, \frac{-\beta+\sqrt{\beta^2-4 \alpha \gamma}}{2 \alpha} \right\}, z_t
  \right]
  \cup
  \left[
  z_{t-1}, \min\left\{z_{t}, \frac{-\beta-\sqrt{\beta^2-4 \alpha \gamma}}{2 \alpha}\right\}
  \right]
 &
  \text{ if }
 \alpha>0,\\
  \left[
z_{t-1}, 
  \min\left\{z_{t}, \frac{-\beta-\sqrt{\beta^2-4 \alpha \gamma}}{2 \alpha} \right\}
  \right]
  \cap
  \left[
   \max\left\{z_{t-1}, \frac{-\beta+\sqrt{\beta^2-4 \alpha \gamma}}{2 \alpha}\right\}
, z_{t}
  \right]
 &
  \text{ if }
 \alpha<0,
  }
 \end{align*}
}
where $\alpha=g_{t,i}^2-g_{t,i'}^2, \beta=2f_{t,i}g_{t,i}-2f_{t,i'}g_{t,i'} $ and $\gamma=f_{t,i}^2-f_{t,i'}^2$. 
 Then, the truncation region $\cZ$ in \eq{eq:cZ} is given as
 \begin{align*}
  \cZ =
  \bigcup_{t=1}^T
  \left(
  \bigcap_{(i, i^\prime) \in \cO(\bm y) \times \left([n] \setminus \cO(\bm y)\right)}
  \cW_{t, (i, i^\prime)}
  \right)
 \end{align*}
\end{lemm}
\noindent
The proof of Lemma~\ref{lemm:topK_outlier_detection} is presented in Appendix A.

\subsection{Discussion: Applicable Class of Robust Regression and Outlier Detection Criteria}
\label{subsec:discussion_applicable}
In the literature of parametric linear programming, if a scalar parameter $\theta \in \RR$ appears in a linear program as
\begin{align}
 \nonumber
 \min_{\bm r}
 ~ 
 &
 (\bm q_0 + \bm q_1 \theta)^\top \bm r
 \\
 \label{eq:general-parametric-LP-form}
 {\rm s.t.}
 \;\;
 &
 S \bm r = \bm u_0 + \bm u_1 \theta, \bm r \ge \bm 0,
\end{align}
then the path of the optimal solutions can be written as piecewise-linear function of the parameter $\theta$. 
It means that, as in LAD regression, if a robust regression estimator is formulated as the solution of a linear program in the above form with $\theta = z$ for a parametrized response vector $\bm y(z) = \bm a + \bm b z, z \in \RR$, both the robust regression estimators and the robust residuals are represented as piecewise-linear functions of $z$ in \eq{eq:piecewise-linear-solution} and \eq{eq:piecewiselinear-residual}, respectively. 
Several regression estimators falls in this framework.
For example, quantile regression~\cite{koenker1978regression,koenker2005quantile} is formulated as a linear program, and the estimators are written as piecewise-linear functions of the quantile order parameter $\tau \in [0, 1]$ (see, e.g., \cite{koenker2005quantile} for the details) by using the above result of parametric linear programming.
Note that there are several variants of the constraint form in \eq{eq:general-parametric-LP-form}, e.g., it could be replaced with $S \bm r \le \bm u_0 + \bm u_1 \theta$ as in \eq{eq:general-parametric-QP-form}.

Similarly, in the literature of parametric quadratic programming, if a scalar parameter $\theta \in \RR$ appears in a quadratic program as
\begin{align}
 \nonumber
 \min_{\bm r}
 ~ 
 &
 \frac{1}{2} \bm r^\top P \bm r + (\bm q_0 + \bm q_1 \theta)^\top \bm r
 \\
 \label{eq:general-parametric-QP-form}
 {\rm s.t.}
 \;\;
 &
 S \bm r \le \bm u_0 + \bm u_1 \theta,
\end{align}
then the path of the optimal solutions can be written as piecewise-linear function of the parameter $\theta$. 
It means that, as in Huber regression, if a robust regression estimator is formulated as the solution of a quadratic program in the above form with $\theta = z$ for a parametrized response vector $\bm y(z) = \bm a + \bm b z, z \in \RR$, both the robust regression estimators and the robust residuals are represented as piecewise-linear functions of $z$ in the forms of \eq{eq:piecewise-linear-solution} and \eq{eq:piecewiselinear-residual}, respectively. 
Note that, when the loss function is defined by using squared error form as in Huber regression, we encounter the quadratic form of the parametrized response vector $\bm y(z)^\top \bm y(z) = \bm a^\top \bm a + 2 \bm a^\top \bm b z + \bm b^\top \bm b z^2$.
Although it seems to be suggesting that quadratic form of the parameters appear in the formulation, since the squared term $\bm b^\top \bm b z^2$ only changes the optimal objective values but does not change the optimal solutions, it is still possible to formulate the squared loss for the parametrized response vectors $\bm y(z) = \bm a + \bm b z$ in the above parametric quadratic programming form. 
In statistics and machine learning, piecewise-linear parametric programming has been used in the context of regularization path (e.g., see \cite{RosZhu07}). 
Regularization path refers to the path of the optimal model parameters for penalized estimators obtained by changing the degrees of penalization.
For example, the regularization path computations for Lasso and SVM (see \cite{osborne2000new,Efron04a} and \cite{hastie2004entire}, respectively) are built within the above parametric quadratic programming framework.
Note that there are several variants of the constraint form in \eq{eq:general-parametric-QP-form}, e.g., it could be replaced with  $S \bm r = \bm u_0 + \bm u_1 \theta, \bm r \ge \bm 0$ as in \eq{eq:general-parametric-LP-form}.

In order to discuss the class of outlier detection criteria which falls within our proposed conditional SI method, let us consider \emph{detection function} $\phi_i: \RR^n \to \RR, i \in [n]$ such that the $i$th instance is detected as outlier if and only if $\phi_i(\bm r^R) \ge 0$. 
In our setting, since the robust residuals linearly change with $z$ within each interval $[z_{t-1}, z_t], t \in [T]$, if sufficiently efficient algorithm for finding the roots of $\phi_i(\bm r^R) = 0, i \in [n]$ within $[z_{t-1}, z_{t}]$ is available, the corresponding outlier detection criterion can be fit into our proposed conditional SI method. 
For the two outlier detection criteria in \S\ref{subsec:outlier_detection_criteria}, the detection functions are given as
\begin{align*}
 \phi_i(\bm r^R) &= |r_i^R| - \xi,
 \\
 \phi_i(\bm r^R) &= |r_i^R| - {\rm top}^k(\{|r_{i^\prime}^R|\}_{i^\prime \in [n]}),
\end{align*}
where ${\rm top}^k(\{|r_{i^\prime}^R|\}_{i^\prime \in [n]})$ is the $k^{\rm th}$ largest value of $|r_{i^\prime}^R|, i^\prime \in [n]$ (although the explicit formulation of the roots of the latter case cannot be simply written out, it is easy to find them by exploiting the fact that each residual monotonically changes within the interval). 
Many other reasonable outlier detection criteria are fit into this framework.
 
\subsection{Relation with Huberized Lasso discussed in \cite{chen2020valid}}
\label{subsec:discussion_on_related_works}

Existing conditional SI studies is almost limited to the cases where the selection event is represented as a set of linear or quadratic inequalities~\cite{lee2016exact,loftus2015selective}. 
However, in many practical problems, over-conditioning with additional extra conditions is needed to characterize the selection events only with linear and/or quadratic inequalities. 
For example, in the seminal paper by Lee et al.~\cite{lee2016exact}, feature selection event of Lasso can be characterized as a set of linear inequalities by extra-conditioning on the signs of the selected features.
Unfortunately, it has been known that over-conditioning leads to the loss of power~\cite{fithian2015selective}.

The closely related work by Chen et al.~\cite{chen2020valid} also falls within this conditional SI framework.
In their work, Conditional SI is discussed for the cases where the outlier detection event can be represented by a set of quadratic inequalities.
The authors showed that simple outlier detection events based on Cook's distance~\cite{cook1977detection} and DFFITS~\cite{welsch1977linear} can be written as a set of quadratic inequalities for which numerical experiments are conducted to demonstrate their performances. 
The authors also discussed that Huber regression can be casted into Lasso-like formulation (called Huberized Lasso~\cite{she2011outlier}) and thus the outlier detection event by Huber regression can be fit into the conditional SI framework by using Lee et al.~\cite{lee2016exact} approach.
However, it should be noted that, since the approach in \cite{lee2016exact} requires over-conditioning by signs, the outlier detection event in Huberized Lasso is also over-conditioned, i.e., it does not coincide with the conditioning defined in \eq{eq:conditional_data_space_y}. 
In other words, the Huberized-Lasso approach discussed in Chen et al.~\cite{chen2020valid} requires over-conditioning, which results in lower detection power than our proposed PLH-based conditional SI method because our proposed method does not require over-conditioning.
The difference between (over-conditioned) Huberized Lasso and our proposed method are clearly demonstrated in the numerical results in \ref{subsec:exp_huberized_lasso} in which the detection power of the former is lower than the latter~\footnote{Since numerical results on Huberized Lasso were not presented in \cite{chen2020valid}, we implemented it by ourself (see Appendix C).}.

\section{Numerical Experiments}
\label{sec:numerical_experiments}

In this section, we demonstrate the performance of the proposed PLH-based conditional SI method on simulated and read data. 

\subsection{Method Options}
In both of simulated and real data experiments, we compared the performances of the following method options.
As options for robust regression estimators, we considered
\begin{itemize}
 \item {\tt LAD}: Least Absolute Deviation (LAD) regression, and
 \item {\tt Huber}: Huber regression.
\end{itemize}
For each of these two robust regression estimators, we considered two outlier detection criteria: 
\begin{itemize}
 \item {\tt Threshold}: Threshold-based outiler detection, and 
 \item {\tt TopK}: Top-$K$-based outiler detection.
\end{itemize}
For each combination of the two robust regression estimators and the two outlier detection criteria, we compared the false positive rates (FPRs) and true positive rates (TPRs) of
\begin{itemize}
 \item {\tt Naive}: Inference without considering the selection bias,
 \item {\tt Bonferroni}: Multiple testing correction by Bonferroni method, and
 \item {\tt PLH-SI}: The proposed conditional SI approach.
\end{itemize}
For Huber regression and threshold-based outlier detection, we also compared our proposed PLH-based conditional SI method with
\begin{itemize}
 \item {\tt HL-SI}: Huberized-Lasso-based conditional SI discussed in \cite{chen2020valid} (see \S\ref{subsec:discussion_on_related_works}) in \S\ref{subsec:exp_huberized_lasso}.
\end{itemize}

\subsection{Experiments on Simulated Data}

\paragraph{Simulated data}
In each trial, we generated a training set for a regression problem $\{(\bm x_i, y_i) \in \RR^p \times \RR \}_{i \in [n]}$ based on the following linear location-shift model
\begin{align*}
 y_i = \beta_0^* + \sum_{j=1}^p \beta_j^* x_{ij} + u_i + \veps_i, i \in [n].
\end{align*}
The design variables $x_{ij}, i \in [n], j \in [p], $ were randomly sampled from $N(0, 1)$ in each trial. 
The true regression coefficients $(\beta_0^*, \beta_1^*, \cdots, \beta_p^*)$ were set to be $(1, 2, 1, 2, \ldots)$.
The noise variance $\sigma^2$ was set to be 1 and assumed to be known {\it a priori}. 
If the location-shift parameter $u_i \neq 0$, we regarded that the $i^{\rm th}$ instance is true outlier. 
The statistical significance level was set as $\alpha = 0.05$.
The number of trials in each experimental setup was 1000. 

\paragraph{False positive rates (FPRs)}
For experiments to check the false positive rates (FPRs), we set the location-shift parameters $u_i = 0$ for all $i \in [n]$. 
As the default experimental setting, we set $n = 20$, $p = 5$, $\delta=1.0$ for {\tt Huber}, $\xi=1$ for {\tt Threshold}, and $K = 1$ for {\tt TopK}.
Experiments were performed in various settings by changing each of these setting parameter from the default value to the following several values:
\begin{itemize}
 \item $n \in \{10, 20, 30, 40, 50\}$,
 \item $p \in \{2, 4, 6, 8, 10\}$,
 \item $\xi \in \{0.8, 1.0, 1.2, 1.4, 1.6\}$, and 
 \item $K \in \{1, 2, 3, 4, 5\}$.
\end{itemize}
In each trial, if multiple outliers are detected, one of them was randomly selected.
On the other hand, if no outliers are detected, we did not count it as a trial, i.e., we repeated the trials until we have 1000 trials in which at least one outlier was detected.
In each experimental setup, the FPRs are estimated by $\sum_{m=1}^{1000} \one\{p_m < \alpha\}$ where $p_m$ is the naive $p$-values for {\tt Naive}, adjusted $p$-values for {\tt Bonferroni} and selective $p$-values in \eq{eq:selective-p-values} for {\tt PLH-SI} at the $m^{\rm th}$ trial, and $\one$ is the index function. 

\begin{figure}[t]
 \begin{center}
  \begin{tabular}{ccc}
   \includegraphics[width=0.33\textwidth]{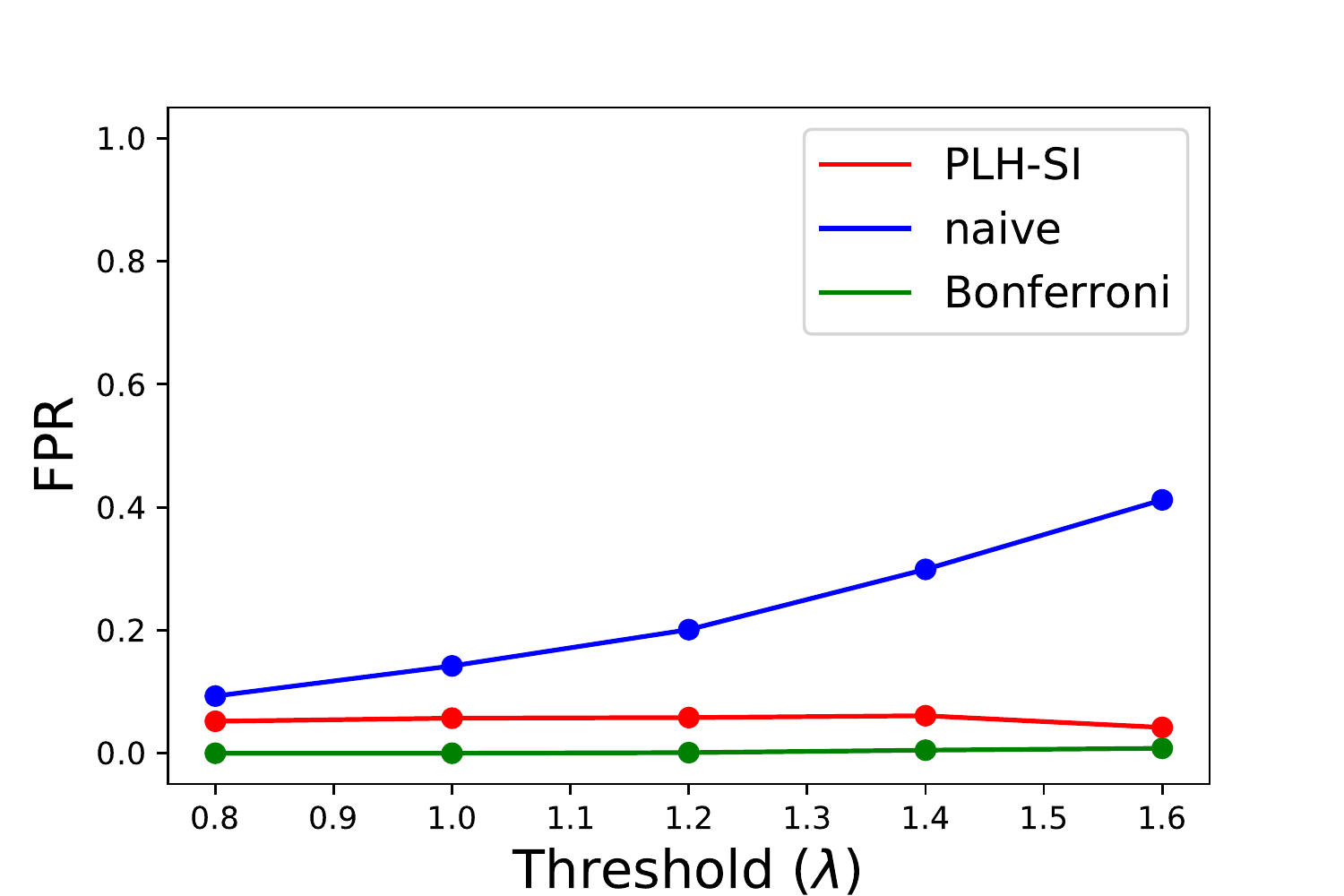}
   &
   \includegraphics[width=0.33\textwidth]{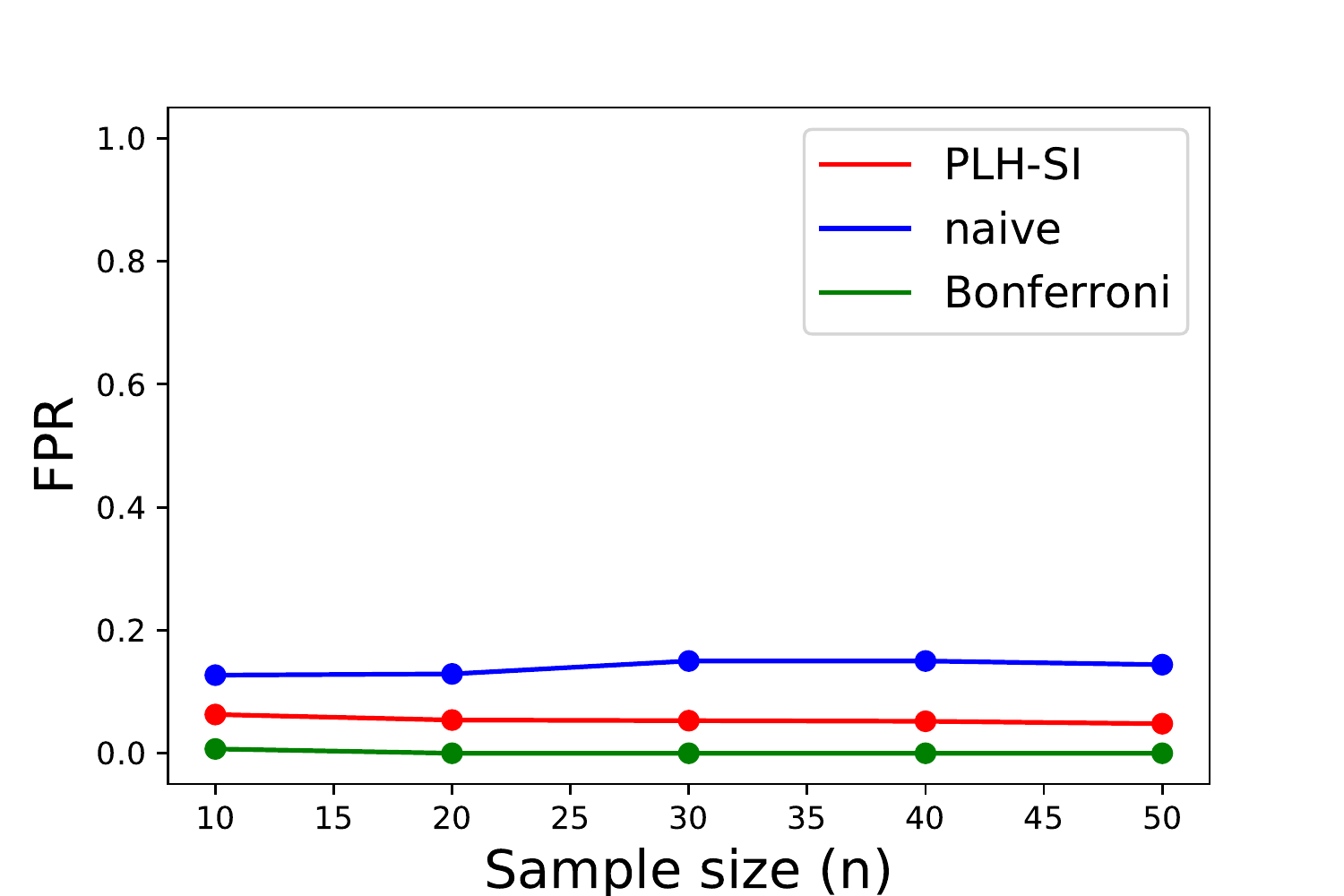}
   &
   \includegraphics[width=0.33\textwidth]{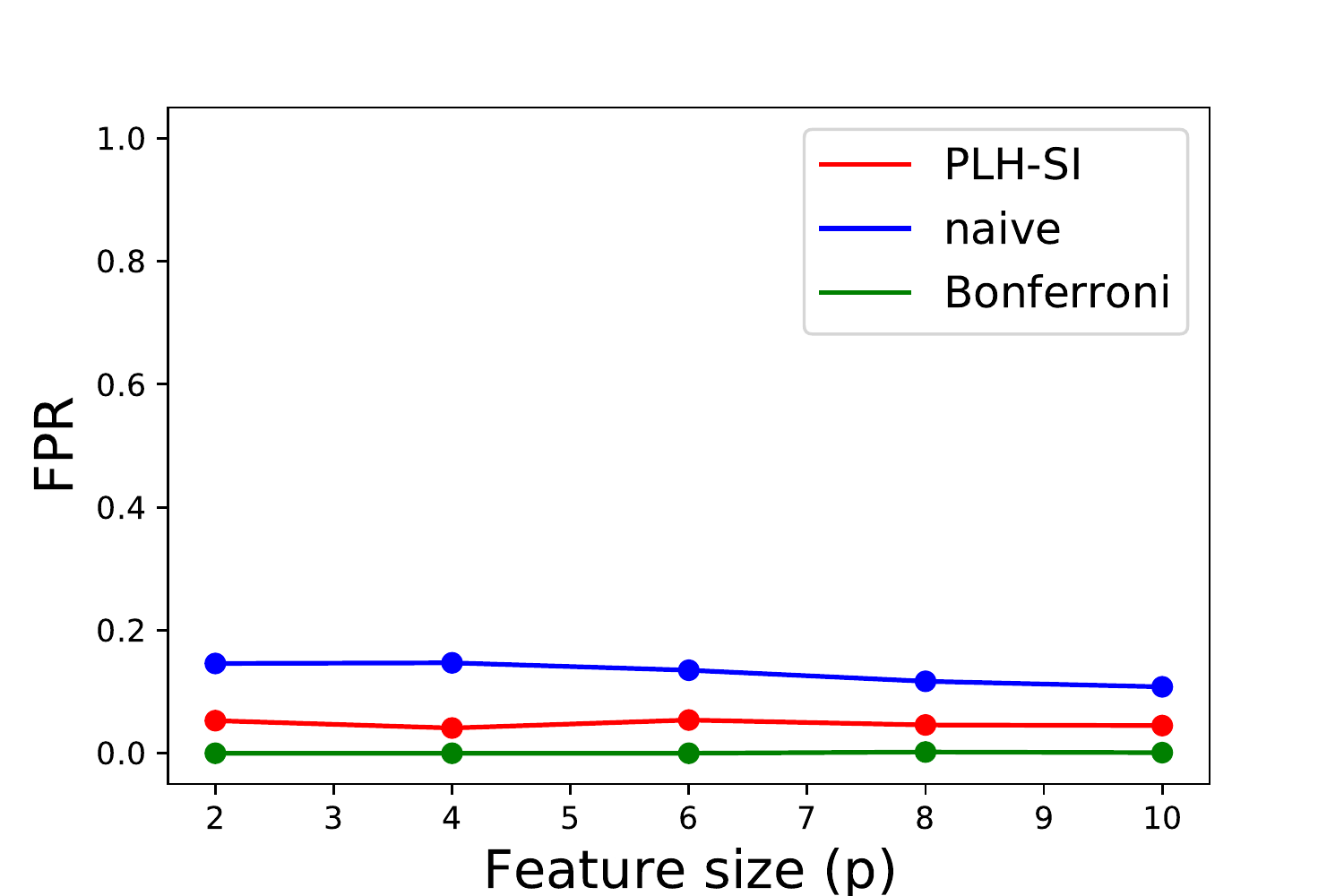}	   
   \\
   \multicolumn{3}{c}{(a) {\tt LAD + Threshold}}
   \\
   \includegraphics[width=0.33\textwidth]{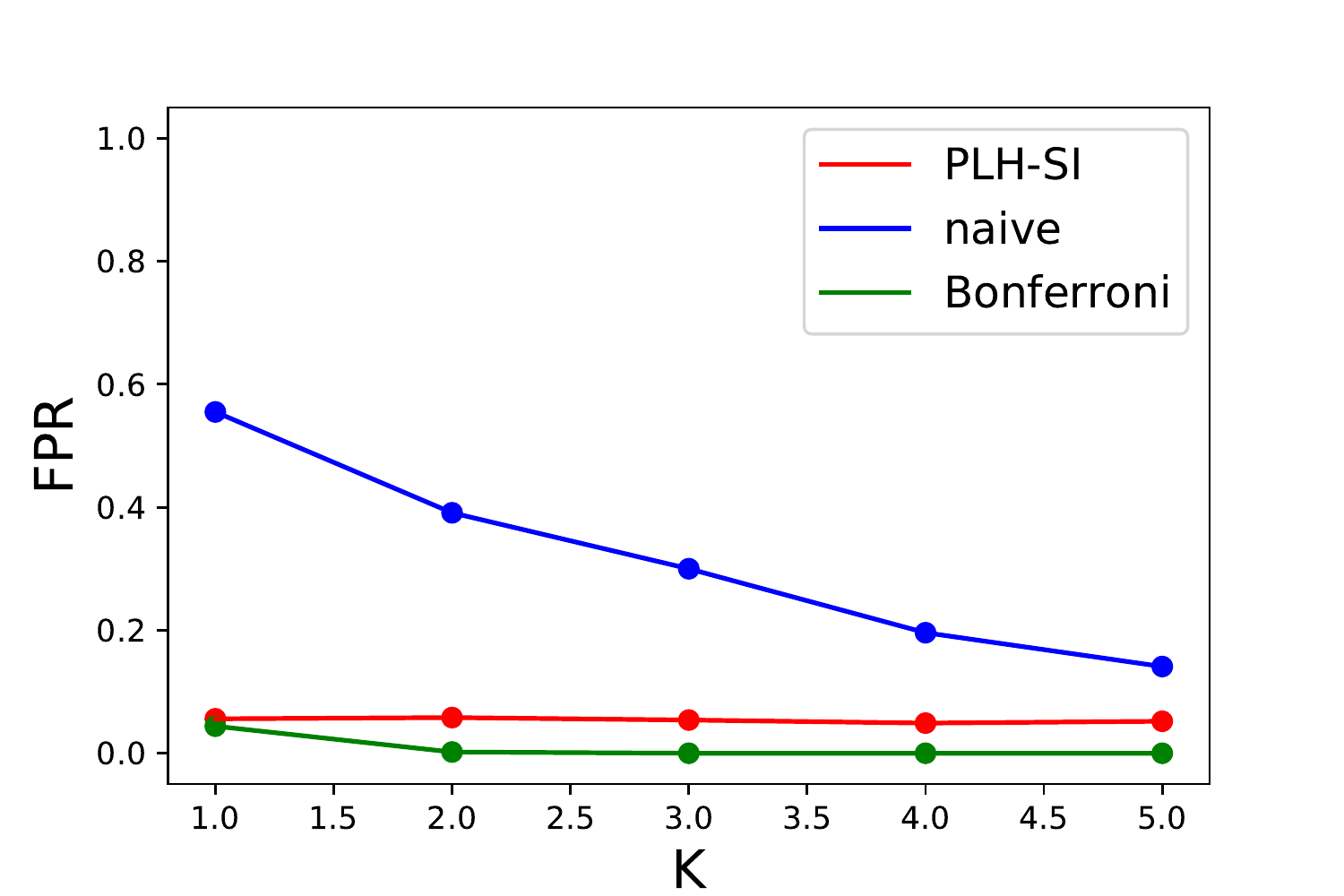}
   &
   \includegraphics[width=0.33\textwidth]{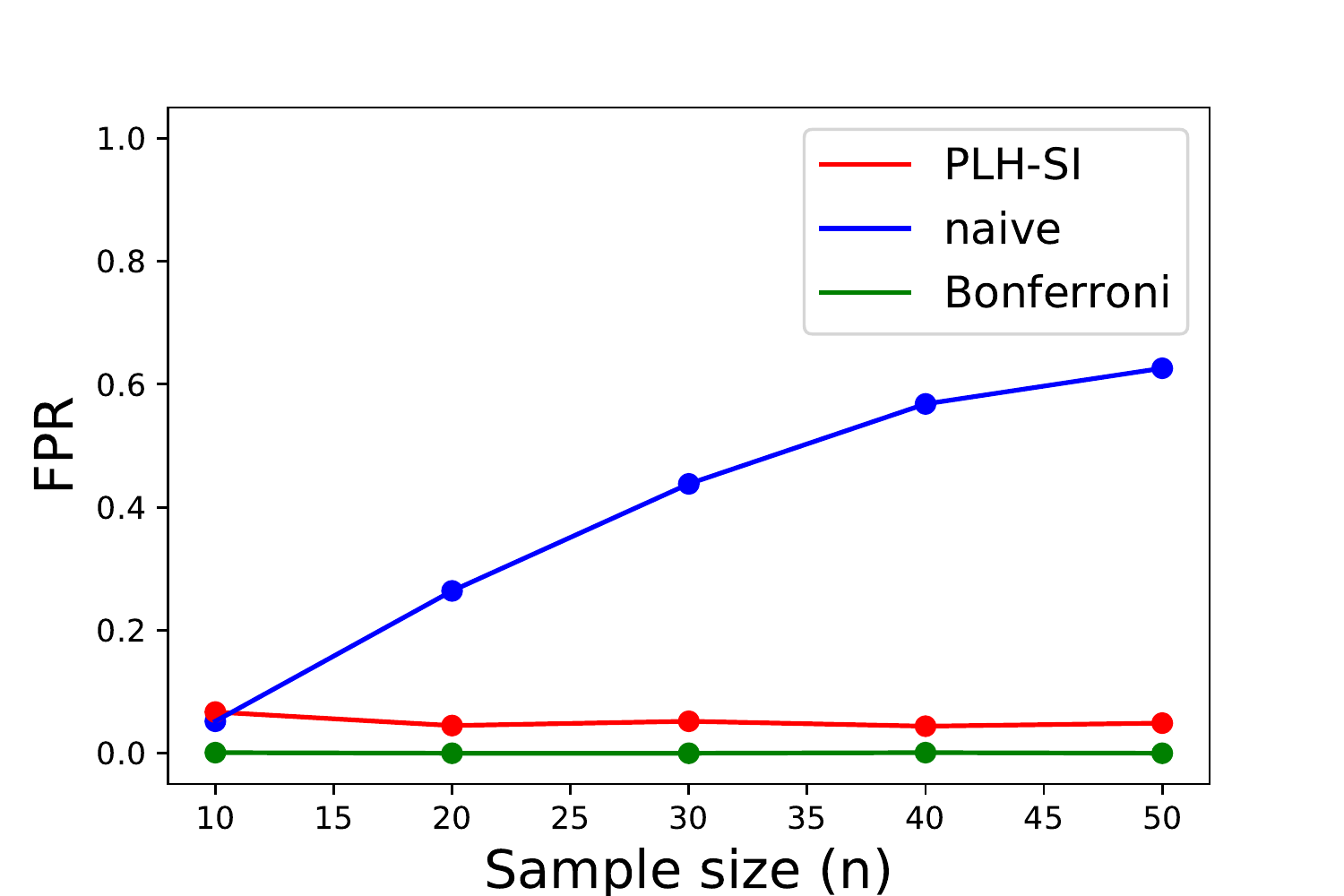}
   &
   \includegraphics[width=0.33\textwidth]{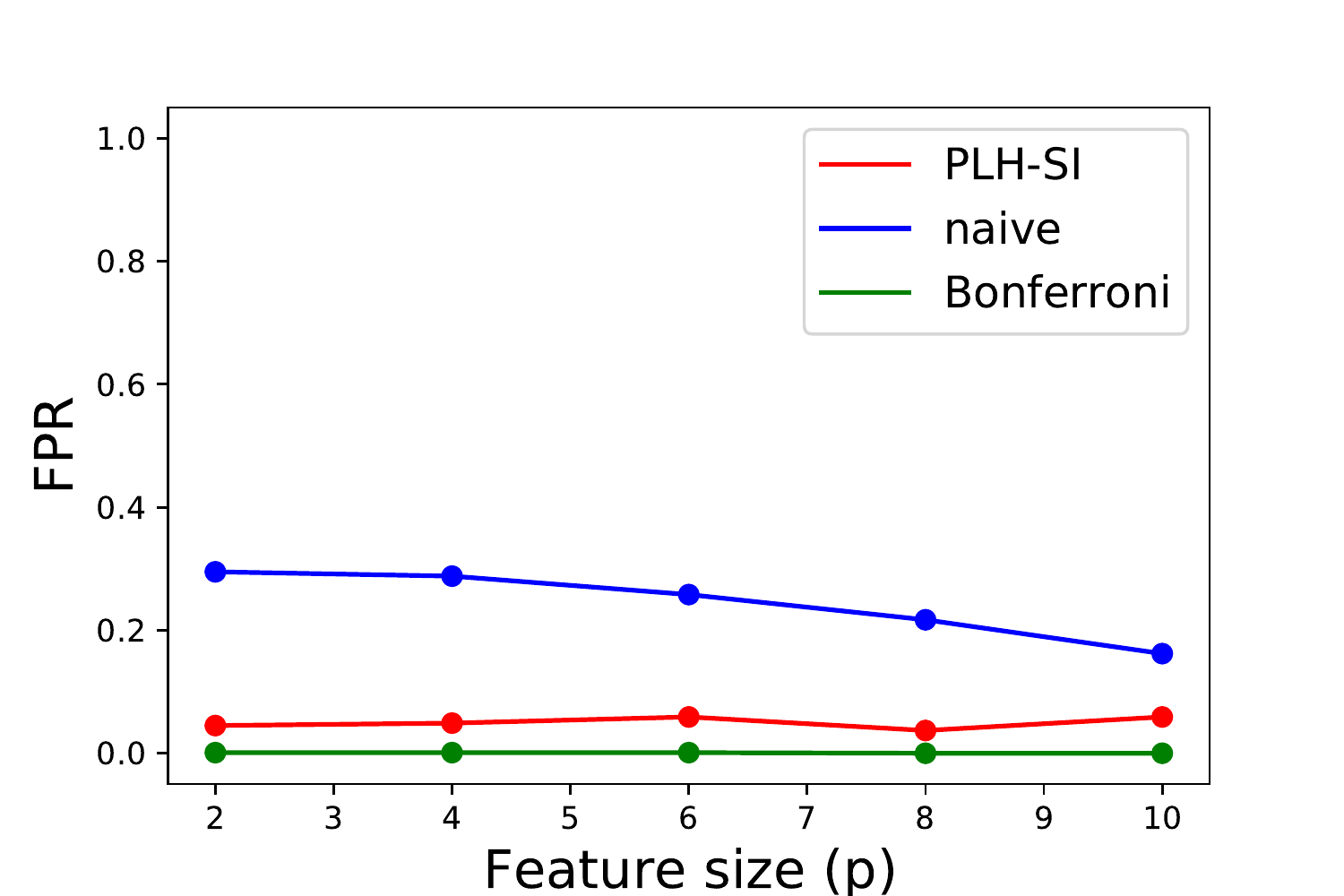}
   \\
   \multicolumn{3}{c}{(b) {\tt LAD + TopK}}
   \\
   \includegraphics[width=0.33\textwidth]{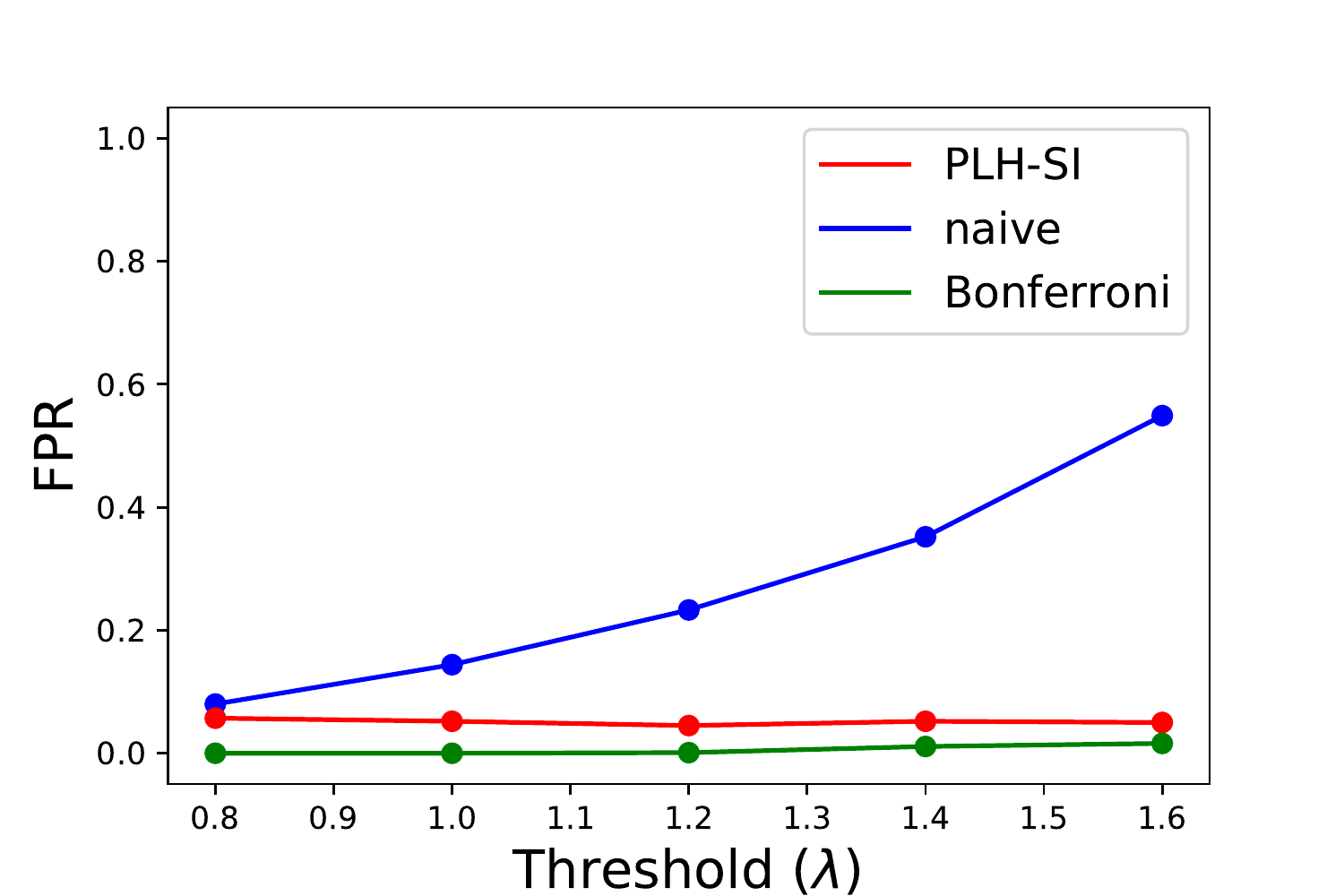}
   &
   \includegraphics[width=0.33\textwidth]{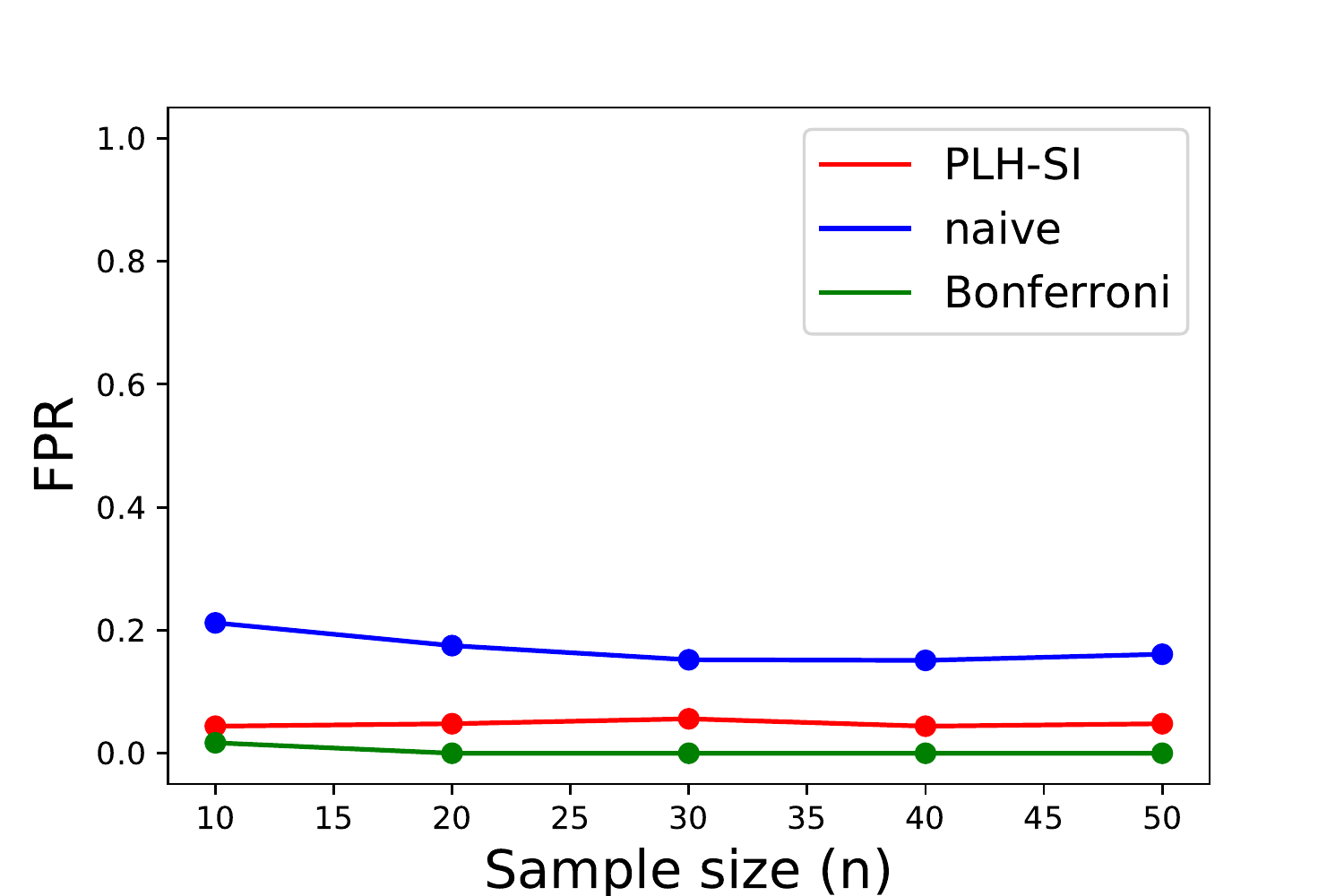}
   &
   \includegraphics[width=0.33\textwidth]{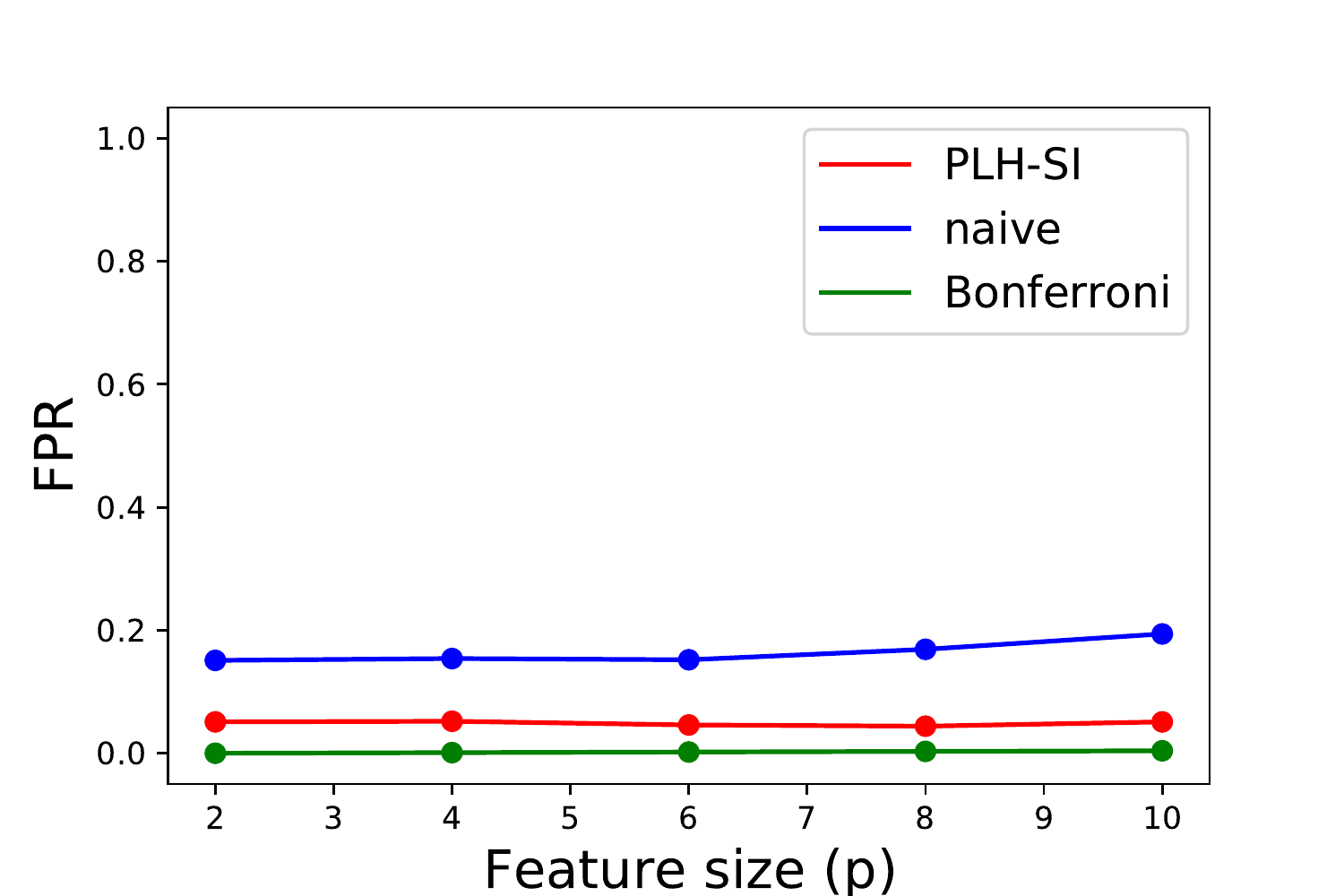}
   \\
   \multicolumn{3}{c}{(c) {\tt Huber + Threshold}}
   \\
   \includegraphics[width=0.33\textwidth]{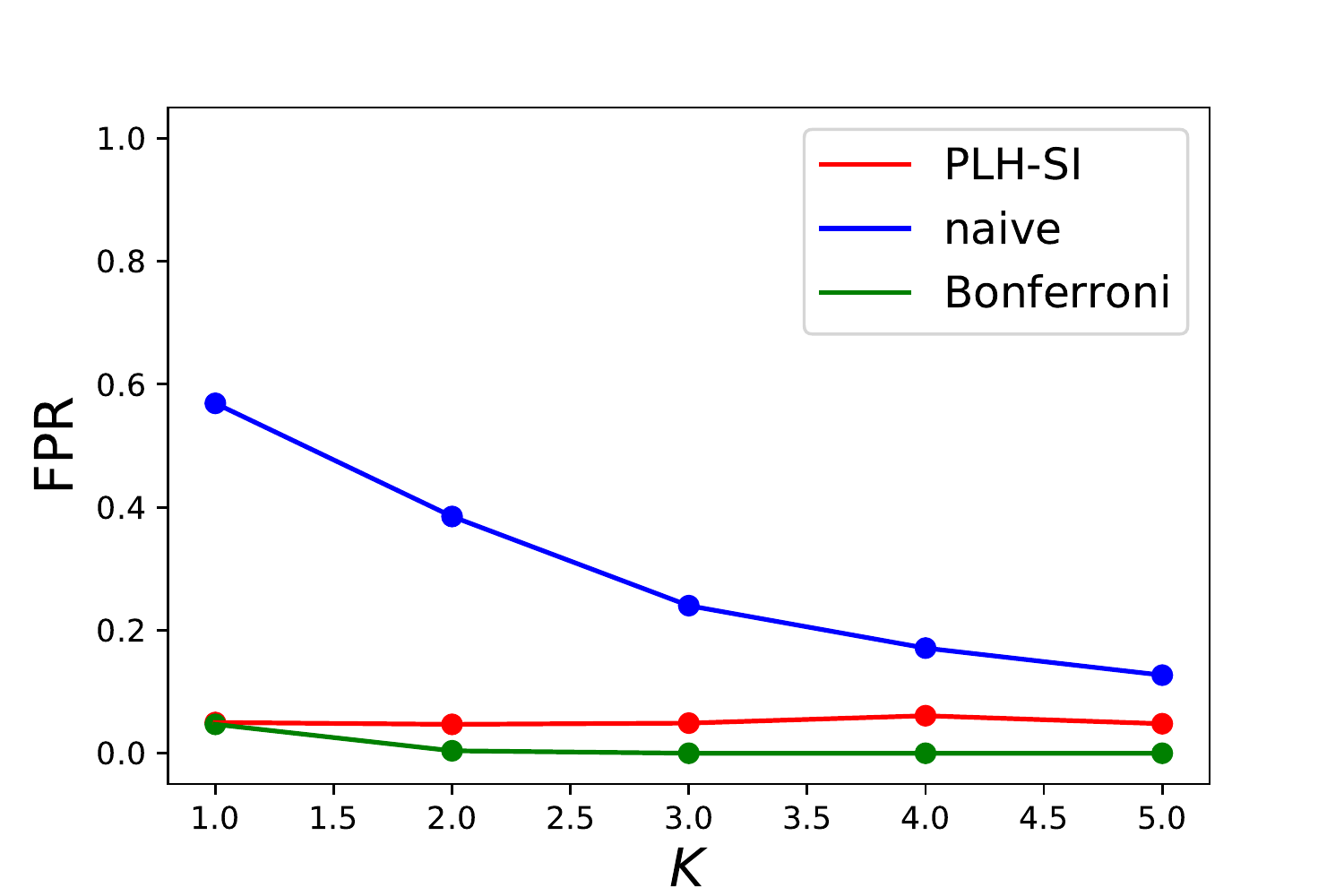}   
   &
   \includegraphics[width=0.33\textwidth]{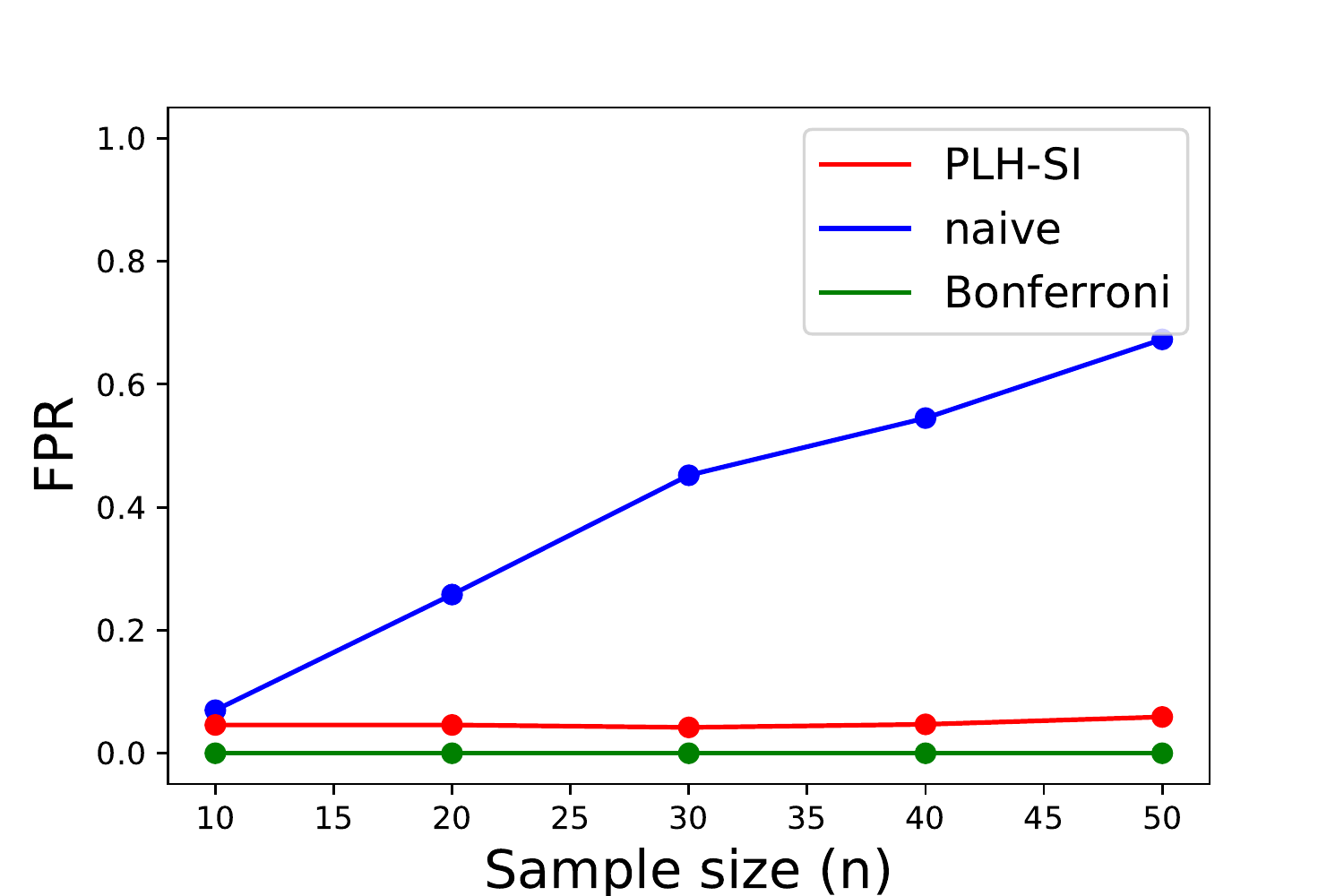}
   &
   \includegraphics[width=0.33\textwidth]{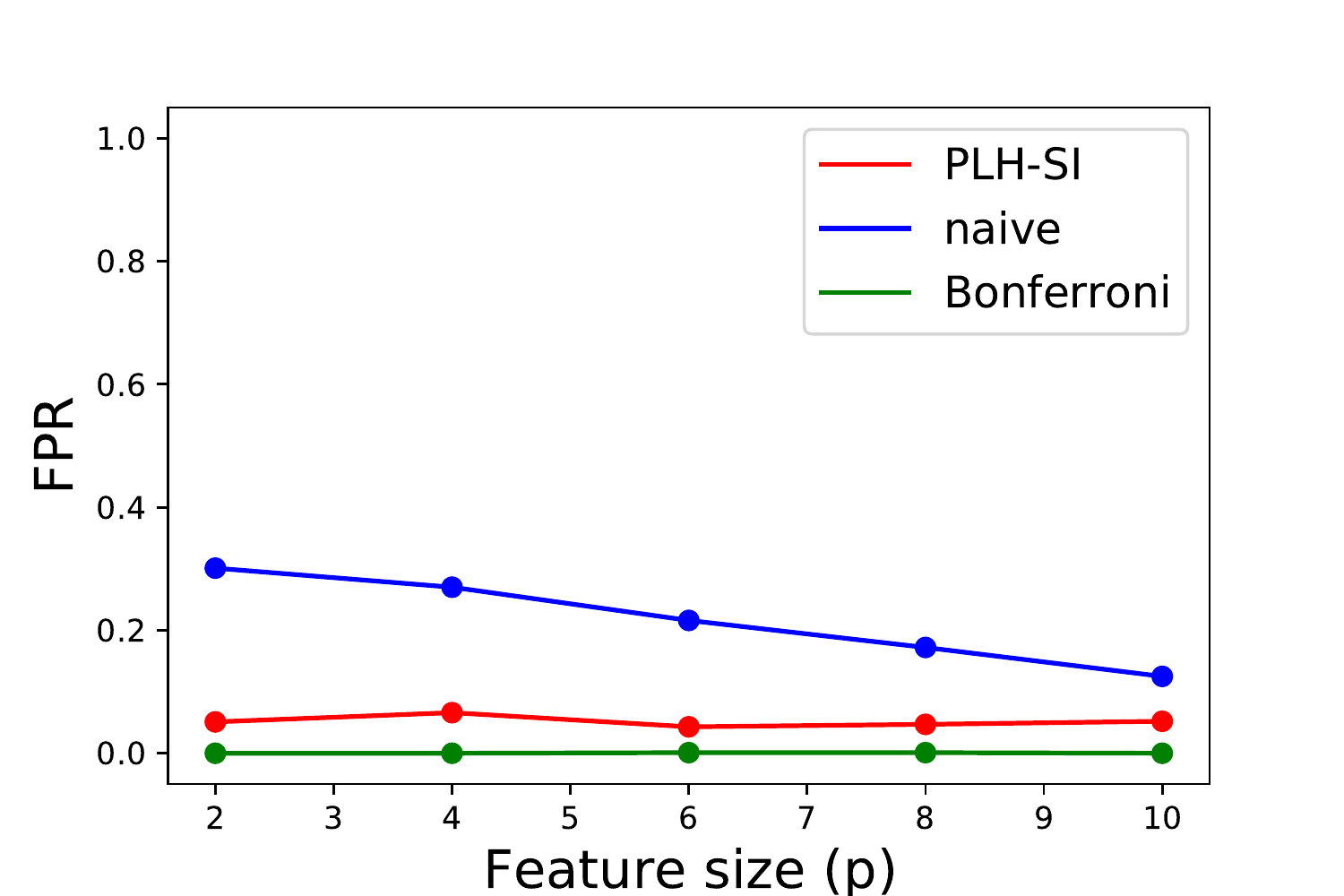}
   \\	
   \multicolumn{3}{c}{(d) {\tt Huber + TopK}}
  \end{tabular}
  \caption{False positive rates of {\tt PLH-SI} (proposed), {\tt Naive} and {\tt Bonferroni} in various setups.}
 \label{fig:fpr}  
 \end{center}
\end{figure}

Figure~\ref{fig:fpr}(a), (b), (c) and (d) show the FPRs of {\tt LAD + Threshold}, {\tt LAD + TopK}, {\tt Huber + Threshold}, and {\tt Huber + TopK}, respectively.
First, it can be seen that {\tt Naive} is not valid in the sense that the FPRs cannot be controlled at the significance level $\alpha = 0.05$.
On the other hand, the proposed {\tt PLH-SI} and {\tt Bonferroni} succeeded in controlling the FPRs at the significance level.
The FPRs of {\tt Bonferroni} was smaller than the significance level $\alpha = 0.05$, which indicates that it is too conservative.
On the other hand, the FPRs of {\tt PLH-SI} is nearly $\alpha = 0.05$ in all settings. 

\paragraph{True positive rates (TPRs)}
For experiments to check the true positive rates (TPRs), we set the 1st instance to be outlier by setting the location-shift parameter as $u_1 \in \{1, 2, 3, 4, 5\}$ and rest of the instances were not regarded as outliers by setting $u_i = 0$ for $i = 2, \ldots, n$. 
In each experimental setup, the TPRs are estimated by counting the fraction that the selective $p$-value of the first instance is smaller than $\alpha$ among the trials in which the 1st instance was detected as an outlier. 
We fixed $n=20$ and $p=5$ and considered $\xi \in \{1, 1.5, 2.0\}$ for {\tt Threshold} and $K \in \{1, 2, 3\}$ for {\tt TopK}. 

\begin{figure}[t]
 \begin{center}
  \begin{tabular}{cc}
   \includegraphics[width=0.33\textwidth]{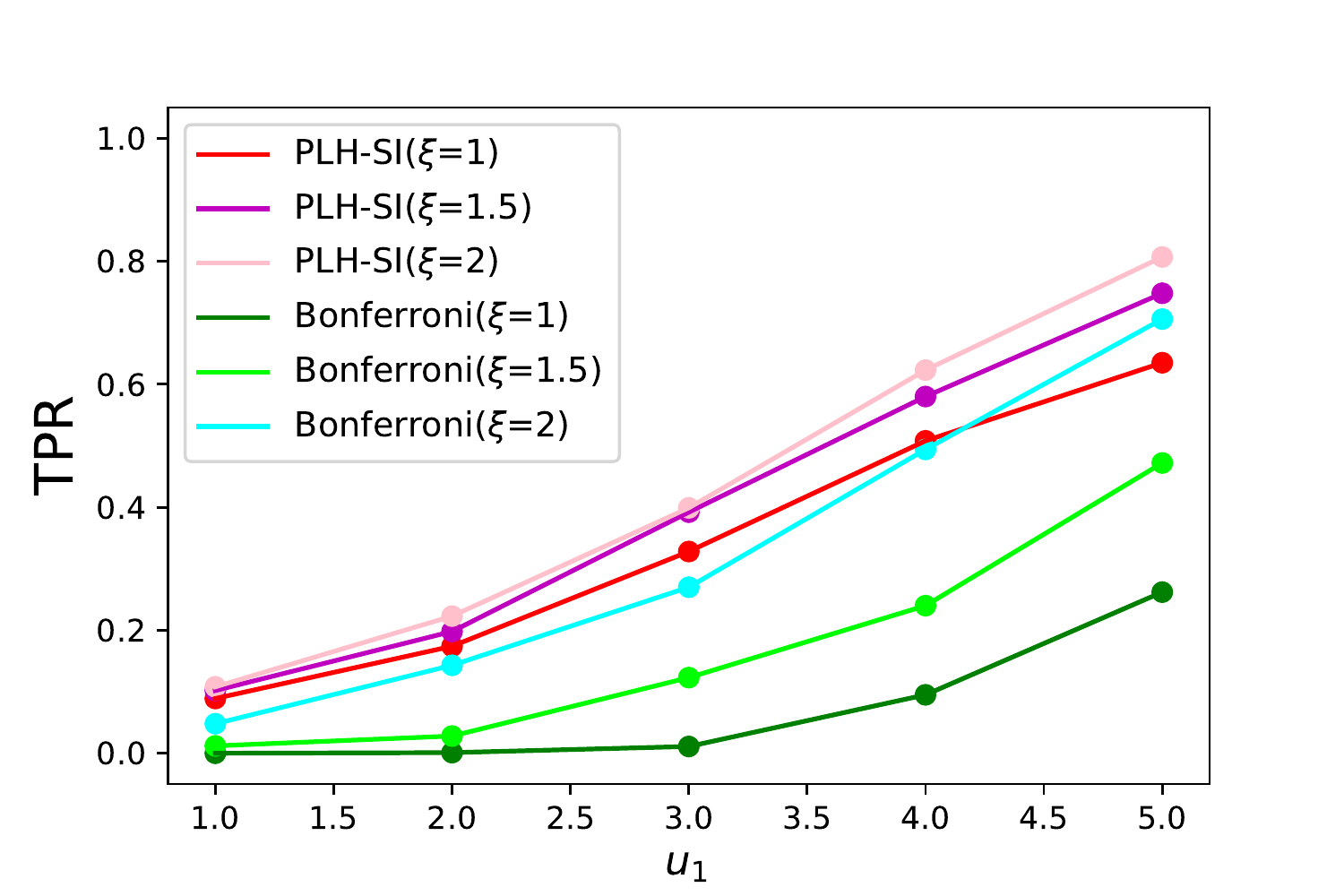}
   &
   \includegraphics[width=0.33\textwidth]{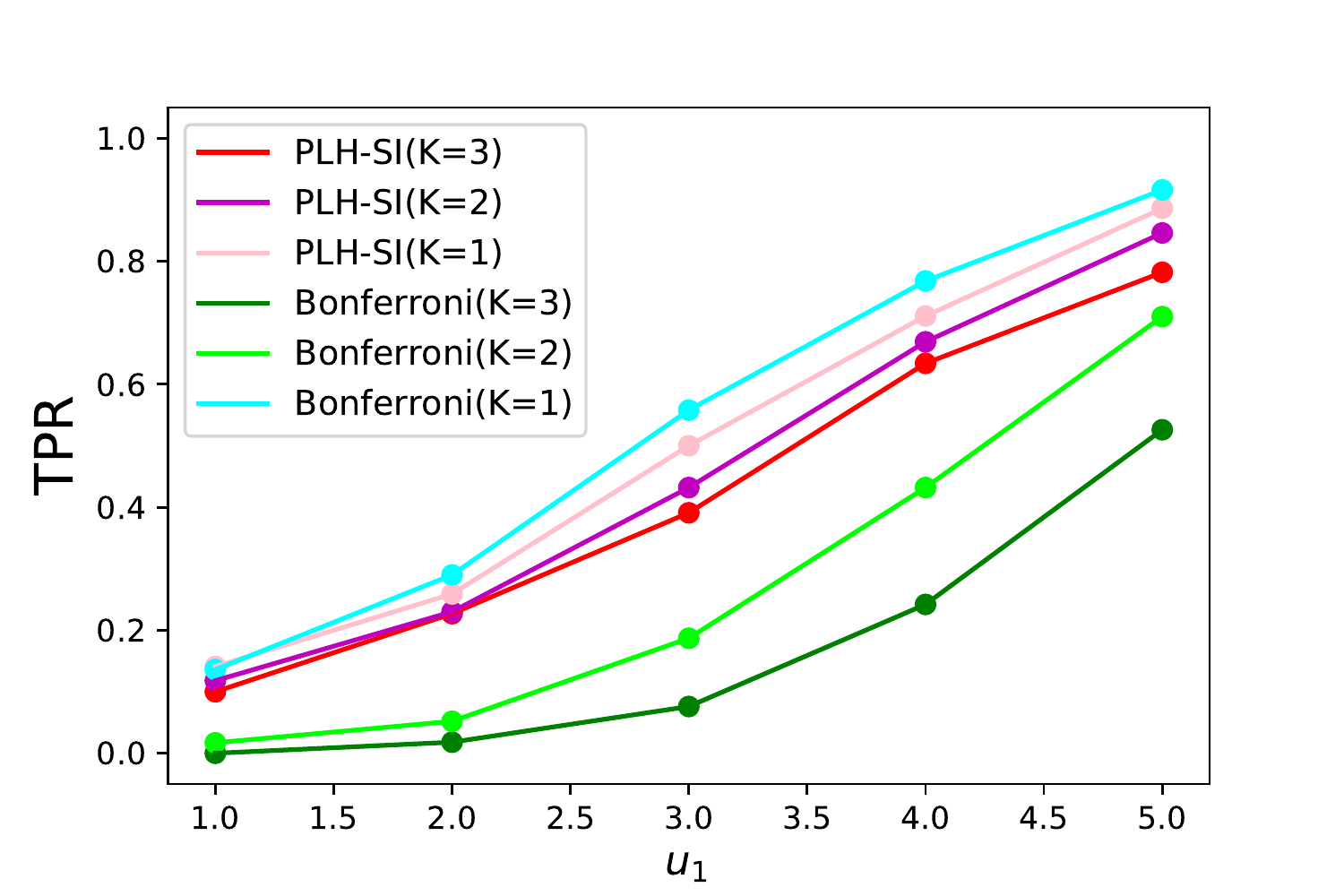} 
   \\
   (a) {\tt LAD + Threshold}
   &
   (b) {\tt LAD + TopK}
   \\
   \includegraphics[width=0.33\textwidth]{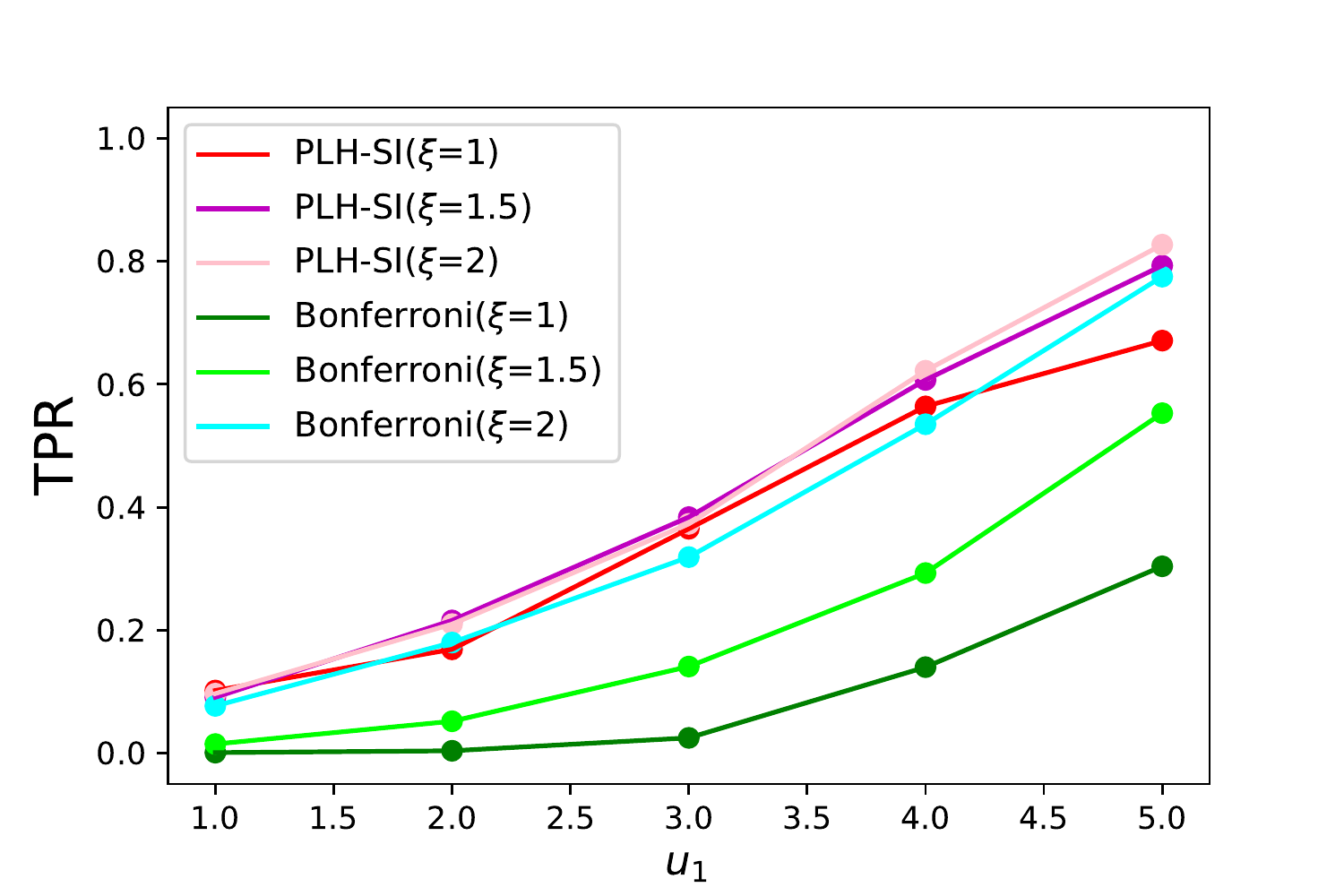} 
   &
   \includegraphics[width=0.33\textwidth]{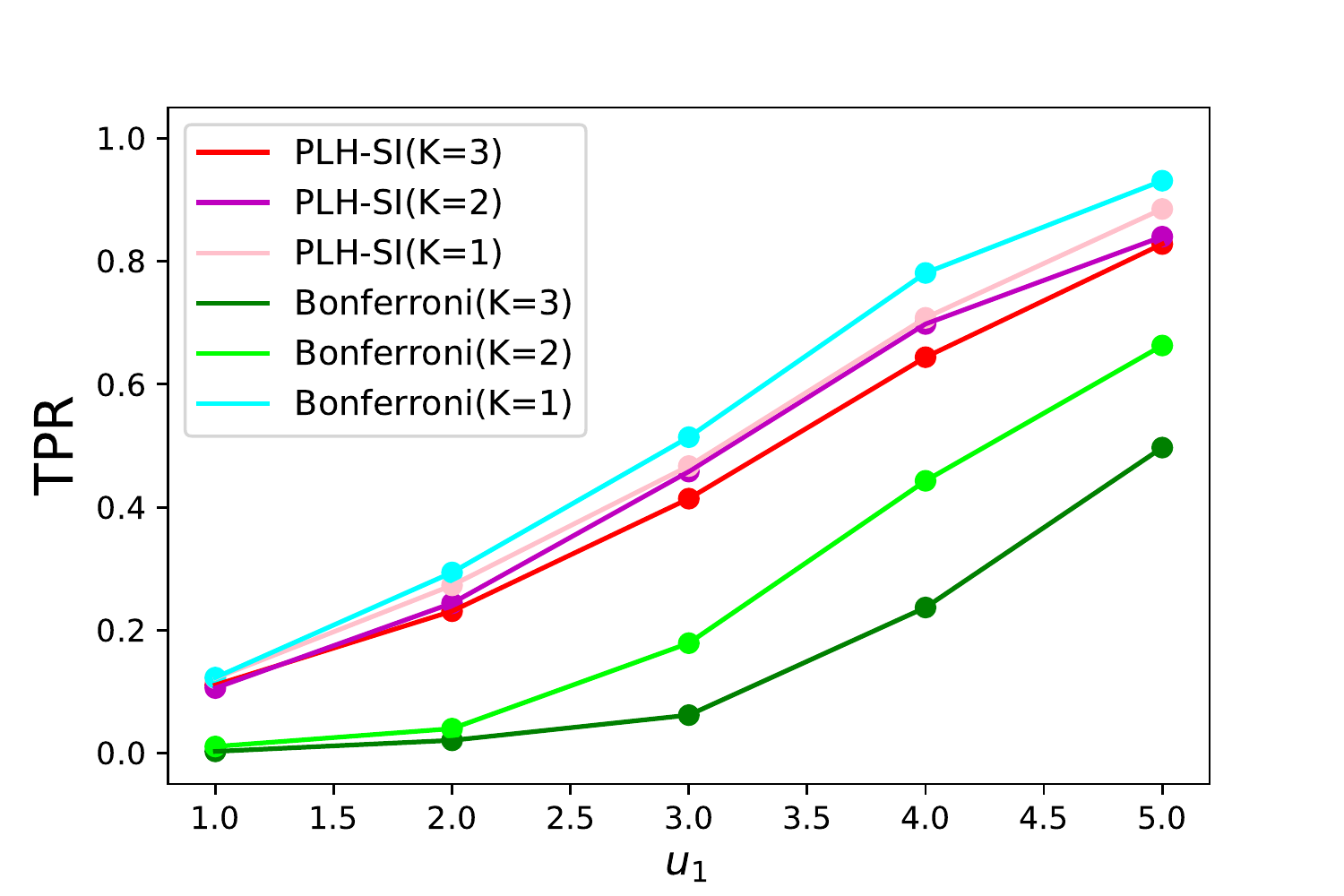}        
   \\
   (c) {\tt Huber + Threshold}
   &
   (d) {\tt Huber + TopK}
  \end{tabular}
  \caption{True positive rates of {\tt PLH-SI} (proposed) and {\tt Bonferroni} in various setups.}
  \label{fig:tpr}
 \end{center}
\end{figure}

Figure~\ref{fig:tpr}(a), (b), (c) and (d) show the TPRs of {\tt LAD + Threshold}, {\tt LAD + TopK}, {\tt Huber + Threshold}, and {\tt Huber + TopK}, respectively.
In these figures, the results for {\tt Naive} are not shown because it cannot control the FPRs.
Comparing {\tt PLH-SI} and {\tt Bonferroni}, it can be confirmed that the detection power of the former is higher than the latter in almost all cases.
The exception is the case of $K=1$ in (b) and (d) where {\tt Bonferroni} has a slightly higher power than {\tt PLH-SI} --- this may be due to the fact that the correction factor of the multiple testing correction by the Bonferroni method for {\tt TopK} at $K=1$ is only $n=20$.
In fact, it can be seen that the power of {\tt Bonferroni} in (b) and (d) decreases as $K$ increases.
In our proposed {\tt PLH-SI}, the power still decreases as $K$ increases, but the degree of decrease is smaller than that of the multiple testing correction by {\tt Bonferroni}.
A similar phenomenon is seen when $\xi$ is varied for {\tt Threshold} in Figs. (a) and (c) --- the detection power decreases for both {\tt Bonferroni} and {\tt PLH-SI} because the number of outliers detected decreases as the value of the threshold $\xi$ becomes smaller, but the amount of the decrease is smaller in our proposed method.

\subsection{Experiments on Real Data}

We demonstrate the performances of the proposed {\tt PLH-SI} on two real datasets which have been analyzed in previous studies in the context of outlier detection \cite{andrews1974robust,hoeting1996method}.
In practice, we should discuss how to determine $\xi$ in {\tt Threshold} and $K$ in {\tt TopK}, but it is non-trivial to incorporate the selection of these tuning parameters into the conditional SI framework, we thus assume that they were predetermined in this paper. 

The first dataset is Brownlee's Stack Loss Plant Data \cite{brownlee1965statistical}.
We denote this dataset as {\it Brownlee}. 
This dataset was studied in plant research on the oxidation of ammonia to nitric acid.
The sample size is $ n = 21 $ and the number of features is $p = 3$.
The three features are {\it Air Flow} (the fow of cooling air), {\it Water Temp} (the cooling water inlet temperature), and {\it Acid Conc} (the concentration of acid), while the response is {\it Stack Loss}.
This dataset has been analyzed as a benchmark dataset in the context of outlier detection \cite{andrews1974robust,hoeting1996method}, and it has been known that the $1^{\rm st}$, $3^{\rm rd}$, $4^{\rm th}$ and $21^{\rm st}$ instances are actually considered as outliers. 
Thus, in the current experiments, we estimated the variance after excluding these known outliers. 

Tables \ref{tab:real1_lad} and \ref{tab:real1_huber} show the $p$-values for {\tt LAD} and {\tt Huber}, respectively, on {\it Brownlee} dataset.
For the results in Table \ref{tab:real1_lad}, we set $\xi = 1.5$ for ${\tt Threshold}$ and $8$ instances are predicted as outliers, and we set $K=8$ for ${\tt TopK}$ so that the same number of outliers are detected for the easiness of comparison. 
In {\tt Naive}, five instances 1, 3, 4, 21, 13 were eventually detected as outliers. 
In {\tt Bonferroni}, only two instances 4, 21 were eventually detected as outliers, and the adjusted $p$-values for other instances exceed 1.0 after correction.
In our proposed {\tt PLH-SI}, four instances 1, 3, 4, 21 were eventually detected as outliers both in {\tt Threshold} and {\tt TopK}. 

For the results in Table \ref{tab:real1_huber}, we set $\xi = 1.5$ for ${\tt Threshold}$ and $8$ instances are predicted as outliers, and we set $K=8$ for ${\tt TopK}$ so that the same number of outliers are detected for the easiness of comparison. 
The detected outliers in all the methods were the same as the case of {\tt LAD}. 
In {\tt Naive}, five instances 1, 3, 4, 21, 13 were eventually detected as outliers. 
In {\tt Bonferroni}, only two instances 4, 21 were eventually detected as outliers, and the adjusted $p$-values for other instances exceed 1.0 after correction.
In our proposed {\tt PLH-SI}, four instances 1, 3, 4, 21 were eventually detected as outliers both in {\tt Threshold} and {\tt TopK}. 

The second dataset is Hill Races Data \cite{atkinson1986influential}.
We denote this dataset as {\it Hill}. 
This dataset is the result of the Scottish hill race.
The sample size is $n = 35$ and the number of features is $p = 2$.
The two features are {\it Dist} (the distance) and {\it Climb} (the total height gained during the route), while the response is {\it Record Time}.
This dataset has been analyzed as a benchmark dataset in the context of outlier detection\cite{andrews1974robust,hoeting1996method}, and it has been known that the $7^{\rm th}$, $18^{\rm th}$ and $33^{\rm rd}$ instances are actually considered as outliers. 
Thus, in the current experiments, we estimated the variance after excluding these known outliers. 

Tables \ref{tab:real2_lad} and \ref{tab:real2_huber} show the $p$-values for {\tt LAD} and {\tt Huber}, respectively, on {\it Hill} dataset.
For the results in Table \ref{tab:real2_lad}, we set $\xi = 6.0$ for ${\tt Threshold}$ and $9$ instances are predicted as outliers, and we set $K=9$ for ${\tt TopK}$ so that the same number of outliers are detected for the easiness of comparison. 
In {\tt Naive}, six instances 7, 18, 33, 6, 14, 19 were eventually detected as outliers. 
In {\tt Bonferroni}, only two instances 7, 18 were eventually detected as outliers, and the adjusted $p$-values for other instances exceed 1.0 after correction.
In our proposed {\tt PLH-SI}, three instances 7, 18, 33 were eventually detected as outliers both in {\tt Threshold} and {\tt TopK}. 

For the results in Table \ref{tab:real2_huber}, we set $\xi = 6.0$ for ${\tt Threshold}$ and $10$ instances are predicted as outliers, and we set $K=10$ for ${\tt TopK}$ so that the same number of outliers are detected for the easiness of comparison. 
In {\tt Naive}, five instances 7, 18, 33, 6, 19 were eventually detected as outliers. 
In {\tt Bonferroni}, only two instance 7, 18 were eventually detected as outliers, and the adjusted $p$-values for other instances exceed 1.0 after correction.
In our proposed {\tt PLH-SI}, three instances 7, 18, 33 were eventually detected as outliers both in {\tt Threshold} and {\tt TopK}. 

In all the results in Tables \ref{tab:real1_lad} to \ref{tab:real2_huber}, as expected, the naive $p$-values are anti-conservative and instances that are not considered to be actual outliers were mistakenly detected as outliers. 
On the other hand, the multiple testing correction by {\tt Bonferroni} was too-conservative and failed to detect some instances that are considered to be actual outliers. 
Our proposed {\tt PLH-SI} selective $p$-values were intermediate between these two methods, suggesting that it can be an appropriate indicator for outlier detection.

\begin{table}[t]
 \begin{center}
  \caption{
  The 8 detected outliers by {\tt LAD} on {\it Brownlee} dataset. 
  The naive $p$-values by {\tt Naive}, adjusted $p$-values by {\tt Bonferroni}, selective $p$-values by {\tt Threshold} (with $\xi = 1.5$) and {\tt TopK} (with $K=8$) are shown. 
  Boldface indicates $p < 0.05$.
  Adjusted $p$-values by {\tt Bonferroni} were set to be 1.00 if it exceeds 1.00 after correction.
  }
  \label{tab:real1_lad}
  \begin{tabular}{|c|c|c|c|c|}
   \hline
   Instance ID & 
   {\tt Naive} &
   {\tt Bonferroni} & 	   
   {\tt Threshold} &
   {\tt TopK} \\
   \hline
   1 &
   $\bm{5.56 \times 10^{-5}}$ &
   $1.00$ &
   $\bm{3.07 \times 10^{-3}}$ &
   $\bm{8.82 \times 10^{-4}}$ \\
   \hline
   3 &
   $\bm{7.31 \times 10^{-6}}$ &
   $1.00$ &
   $\bm{6.21 \times 10^{-4}}$ &
   $\bm{1.29 \times 10^{-4}}$ \\
   \hline
   4 &
   $\bm{7.43 \times 10^{-12}}$ &
   $\bm{1.51 \times 10^{-6}}$ &
   $\bm{5.04 \times 10^{-5}}$ &
   $\bm{3.44 \times 10^{-6}}$ \\
   \hline
   21 &
   $\bm{4.23 \times 10^{-12}}$ &
   $\bm{8.60 \times 10^{-7}}$ &
   $\bm{5.69 \times 10^{-4}}$ &
   $\bm{2.38 \times 10^{-4}}$ \\
   \hline
   6 &
   $2.44 \times 10^{-1}$ &
   $1.00$ &
   $9.38 \times 10^{-1}$ &
   $9.75 \times 10^{-1}$ \\
   \hline
   13 &
   $\bm{1.16 \times 10^{-2}}$ &
   $1.00$ &
   $1.37 \times 10^{-1}$ &
   $8.81 \times 10^{-2}$ \\
   \hline
   14 &
   $1.04 \times 10^{-1}$ &
   $1.00$ &
   $4.56 \times 10^{-1}$ &
   $4.24 \times 10^{-1}$ \\
   \hline
   20 &
   $1.26 \times 10^{-1}$ &
   $1.00$ &
   $6.63 \times 10^{-1}$ &
   $6.07 \times 10^{-1}$ \\
   \hline
  \end{tabular}
 \end{center}
 \end{table}
  
\begin{table}[t]
\begin{center}
  \caption{The 8 detected outliers by {\tt Huber} on {\it Brownlee} dataset (see the caption of Table \ref{tab:real1_lad} for the details).}
  \label{tab:real1_huber}
  \begin{tabular}{|c|c|c|c|c|}
   \hline
   Instance ID & 
   {\tt Naive} &
   {\tt Bonferroni} & 	   
   {\tt Threshold} &
   {\tt TopK} \\
   \hline
   1 &
   $\bm{1.91 \times 10^{-4}}$ &
   $1.00$ &
   $\bm{2.83 \times 10^{-3}}$ &
   $\bm{2.56 \times 10^{-3}}$ \\
   \hline
   3 &
   $\bm{1.03 \times 10^{-5}}$ &
   $1.00$ &
   $\bm{8.27 \times 10^{-5}}$ &
   $\bm{6.30 \times 10^{-5}}$ \\
   \hline
   4 &
   $\bm{7.02 \times 10^{-12}}$ &
   $\bm{1.43 \times 10^{-6}}$ &
   $\bm{4.43 \times 10^{-7}}$ &
   $\bm{3.88 \times 10^{-11}}$ \\
   \hline
   21 &
   $\bm{1.40 \times 10^{-11}}$ &
   $\bm{2.85 \times 10^{-6}}$ &
   $\bm{4.13 \times 10^{-10}}$ &
   $\bm{2.06 \times 10^{-10}}$ \\
   \hline
   6 &
   $2.73 \times 10^{-1}$ &
   $1.00$ &
   $5.97 \times 10^{-1}$ &
   $7.30 \times 10^{-1}$ \\
   \hline
   13 &
   $\bm{1.76 \times 10^{-2}}$ &
   $1.00$ &
   $1.17 \times 10^{-1}$ &
   $1.08 \times 10^{-1}$ \\
   \hline
   15 &
   $1.65 \times 10^{-1}$ &
   $1.00$ &
   $8.98 \times 10^{-1}$ &
   $9.96 \times 10^{-1}$ \\
   \hline
   20 &
   $1.10 \times 10^{-1}$ &
   $1.00$ &
   $6.87 \times 10^{-1}$ &
   $5.20 \times 10^{-1}$ \\
   \hline
  \end{tabular}
 \end{center}
\end{table}

\begin{table}[t]
 \begin{center}
  \caption{
  The 9 detected outliers by {\tt LAD} on {\it Hill} dataset (see the caption of Table \ref{tab:real1_lad} for the details). 
  }
  \label{tab:real2_lad}
  \begin{tabular}{|c|c|c|c|c|}
   \hline
   Instance ID & 
   {\tt Naive} &
   {\tt Bonferroni} & 	   
   {\tt Threshold} &
   {\tt TopK} \\
   \hline
   7 &
   $\bm{6.77 \times 10^{-19}}$ &
   $\bm{4.78 \times 10^{-11}}$ &
   $\bm{1.91 \times 10^{-5}}$ &
   $\bm{1.61 \times 10^{-9}}$ \\
   \hline
   18 &
   $\bm{2.15 \times 10^{-43}}$ &
   $\bm{1.52 \times 10^{-35}}$ &
   $\bm{1.34 \times 10^{-25}}$ &
   $\bm{1.76 \times 10^{-32}}$ \\
   \hline
   33 &
   $\bm{2.43 \times 10^{-6}}$ &
   $1.00$ &
   $\bm{2.52 \times 10^{-4}}$ &
   $\bm{3.95 \times 10^{-5}}$ \\
   \hline
   6 &
   $\bm{3.76 \times 10^{-2}}$ &
   $1.00$ &
   $1.72 \times 10^{-1}$ &
   $1.42 \times 10^{-1}$ \\
   \hline
   14 &
   $\bm{4.94 \times 10^{-2}}$ &
   $1.00$ &
   $3.90 \times 10^{-1}$ &
   $3.16 \times 10^{-1}$ \\
   \hline
   16 &
   $2.53 \times 10^{-1}$ &
   $1.00$ &
   $3.32 \times 10^{-1}$ &
   $6.33 \times 10^{-1}$ \\
   \hline
   19 &
   $\bm{1.98 \times 10^{-2}}$ &
   $1.00$ &
   $2.33 \times 10^{-1}$ &
   $1.85 \times 10^{-1}$ \\
   \hline
   24 &
   $1.28 \times 10^{-1}$ &
   $1.00$ &
   $6.55 \times 10^{-1}$ &
   $5.06 \times 10^{-1}$ \\
   \hline
   30 &
   $1.15 \times 10^{-1}$ &
   $1.00$ &
   $4.85 \times 10^{-1}$ &
   $6.38 \times 10^{-1}$ \\
   \hline
  \end{tabular}
 \end{center}
\end{table}

\begin{table}[t]
 \begin{center}
  \caption{
  The 10 detected outliers by {\tt Huber} on {\it Hill} dataset (see the caption of Table \ref{tab:real1_lad} for the details). 
  }
  \label{tab:real2_huber}
  \begin{tabular}{|c|c|c|c|c|}
   \hline
   Instance ID & 
   {\tt Naive} &
   {\tt Bonferroni} & 	   
   {\tt Threshold} &
   {\tt TopK} \\
   \hline
   7 &
   $\bm{3.69 \times 10^{-18}}$ &
   $\bm{6.77 \times 10^{-10}}$ &
   $\bm{3.49 \times 10^{-7}}$ &
   $\bm{3.35 \times 10^{-10}}$ \\
   \hline
   18 &
   $\bm{4.02 \times 10^{-43}}$ &
   $\bm{7.38 \times 10^{-35}}$ &
   $\bm{1.39 \times 10^{-16}}$ &
   $\bm{2.22 \times 10^{-27}}$ \\
   \hline
   33 &
   $\bm{3.58 \times 10^{-6}}$ &
   $1.00$ &
   $\bm{7.19 \times 10^{-5}}$ &
   $\bm{1.39 \times 10^{-5}}$ \\
   \hline
   6 &
   $\bm{4.68 \times 10^{-2}}$ &
   $1.00$ &
   $1.51 \times 10^{-1}$ &
   $1.68 \times 10^{-1}$ \\
   \hline
   14 &
   $5.12 \times 10^{-2}$ &
   $1.00$ &
   $2.79 \times 10^{-1}$ &
   $2.01 \times 10^{-1}$ \\
   \hline
   16 &
   $2.25 \times 10^{-1}$ &
   $1.00$ &
   $6.48 \times 10^{-1}$ &
   $9.44 \times 10^{-1}$ \\
   \hline
   19 &
   $\bm{1.74 \times 10^{-2}}$ &
   $1.00$ &
   $4.40 \times 10^{-1}$ &
   $7.62 \times 10^{-2}$ \\
   \hline
   24 &
   $1.46 \times 10^{-1}$ &
   $1.00$ &
   $5.85 \times 10^{-1}$ &
   $5.74 \times 10^{-1}$ \\
   \hline
   26 &
   $2.47 \times 10^{-1}$ &
   $1.00$ &
   $8.38 \times 10^{-2}$ &
   $5.75 \times 10^{-1}$ \\
   \hline
   30 &
   $1.09 \times 10^{-1}$ &
   $1.00$ &
   $6.44 \times 10^{-1}$ &
   $4.66 \times 10^{-1}$ \\
   \hline
  \end{tabular}
 \end{center}
\end{table}

\subsection{Comparison with Over-Conditioned Huberized Lasso}
\label{subsec:exp_huberized_lasso}
\begin{figure}[t]
 \begin{center}
  \begin{tabular}{cc}
   \includegraphics[width=0.33\textwidth]{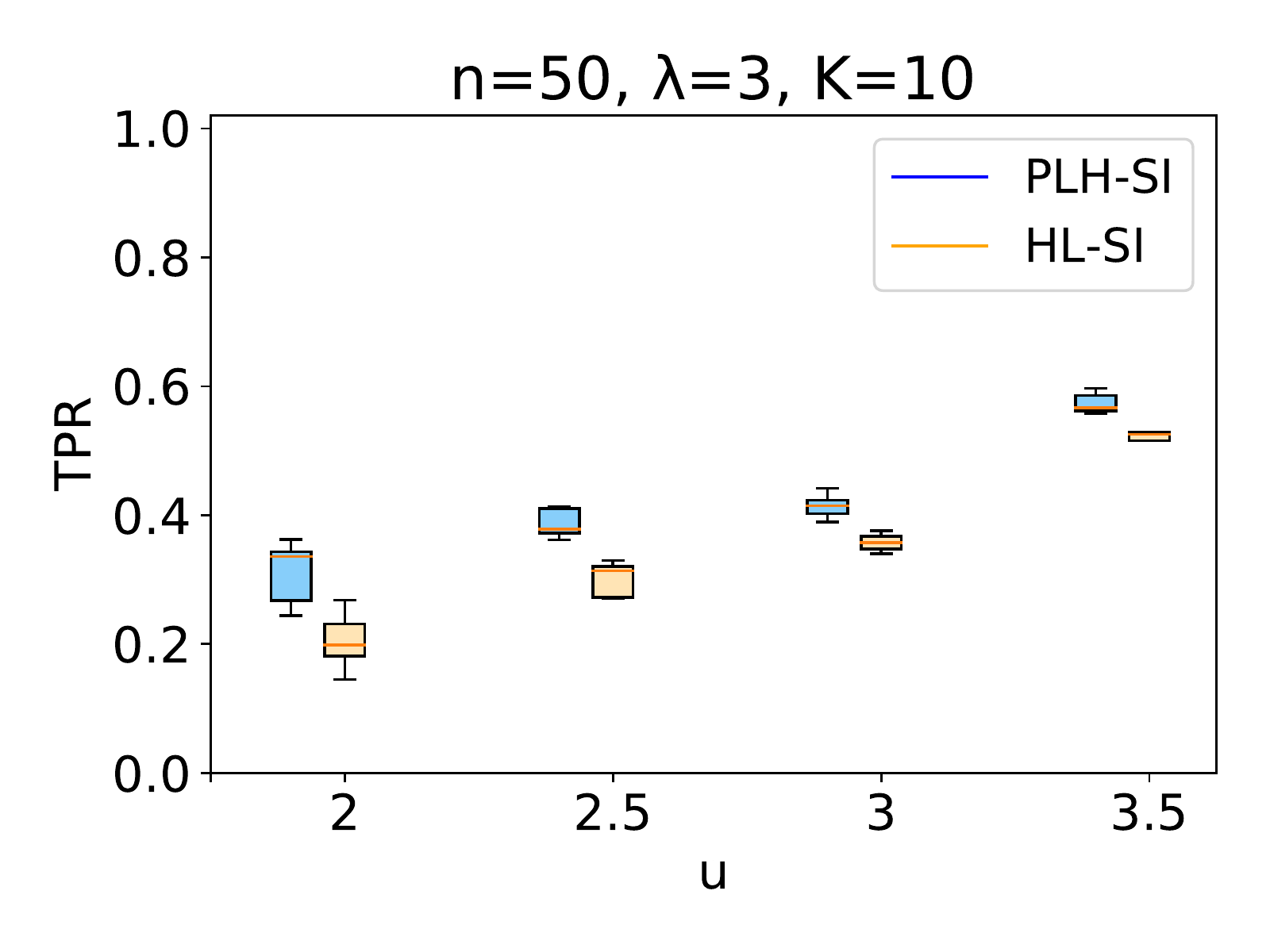}
   &
   \includegraphics[width=0.33\textwidth]{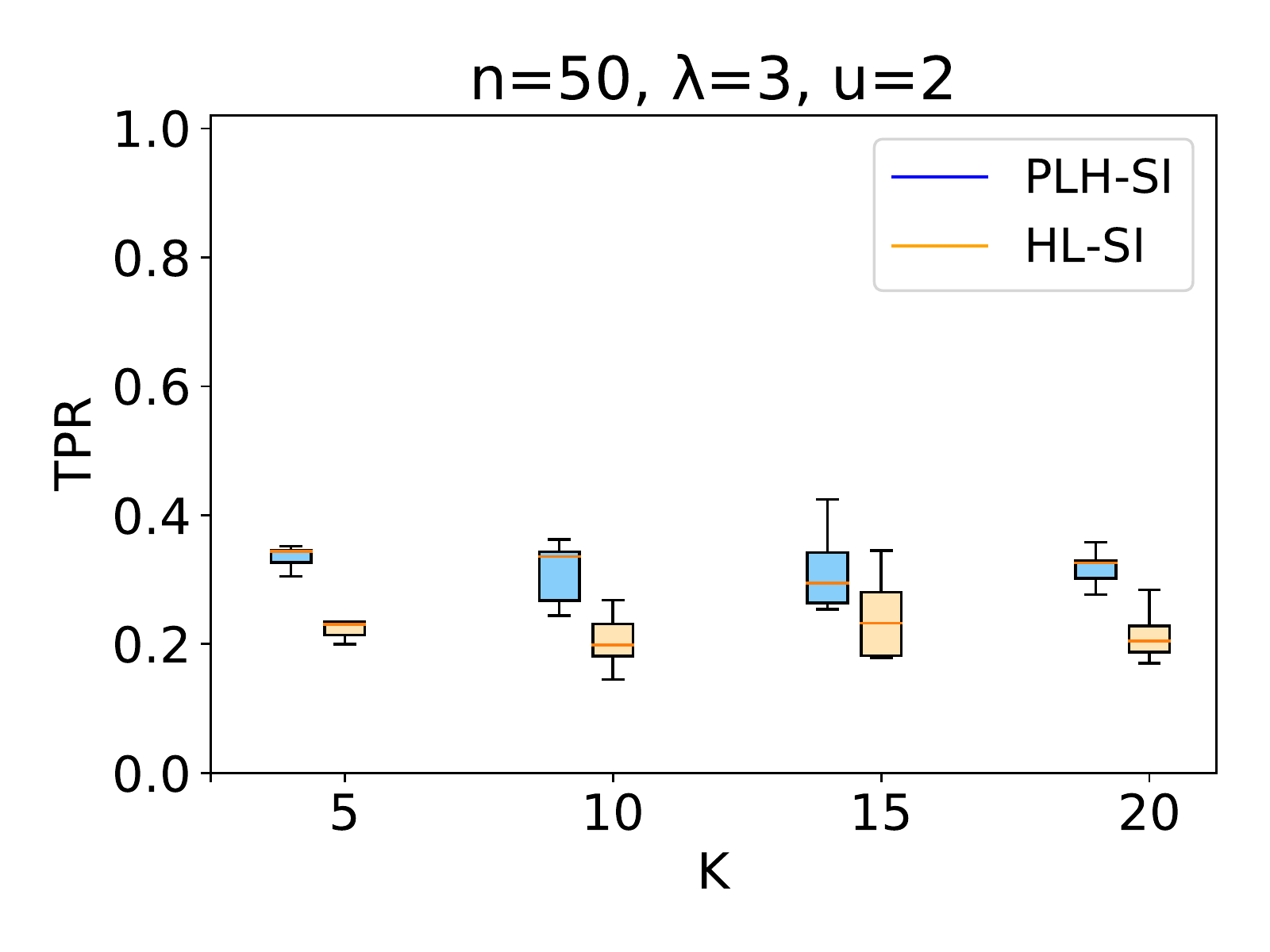}
   \\
   (a) Comparison for different $u$s
   &
   (b) Comparison for different $K$s
   \\
   \includegraphics[width=0.33\textwidth]{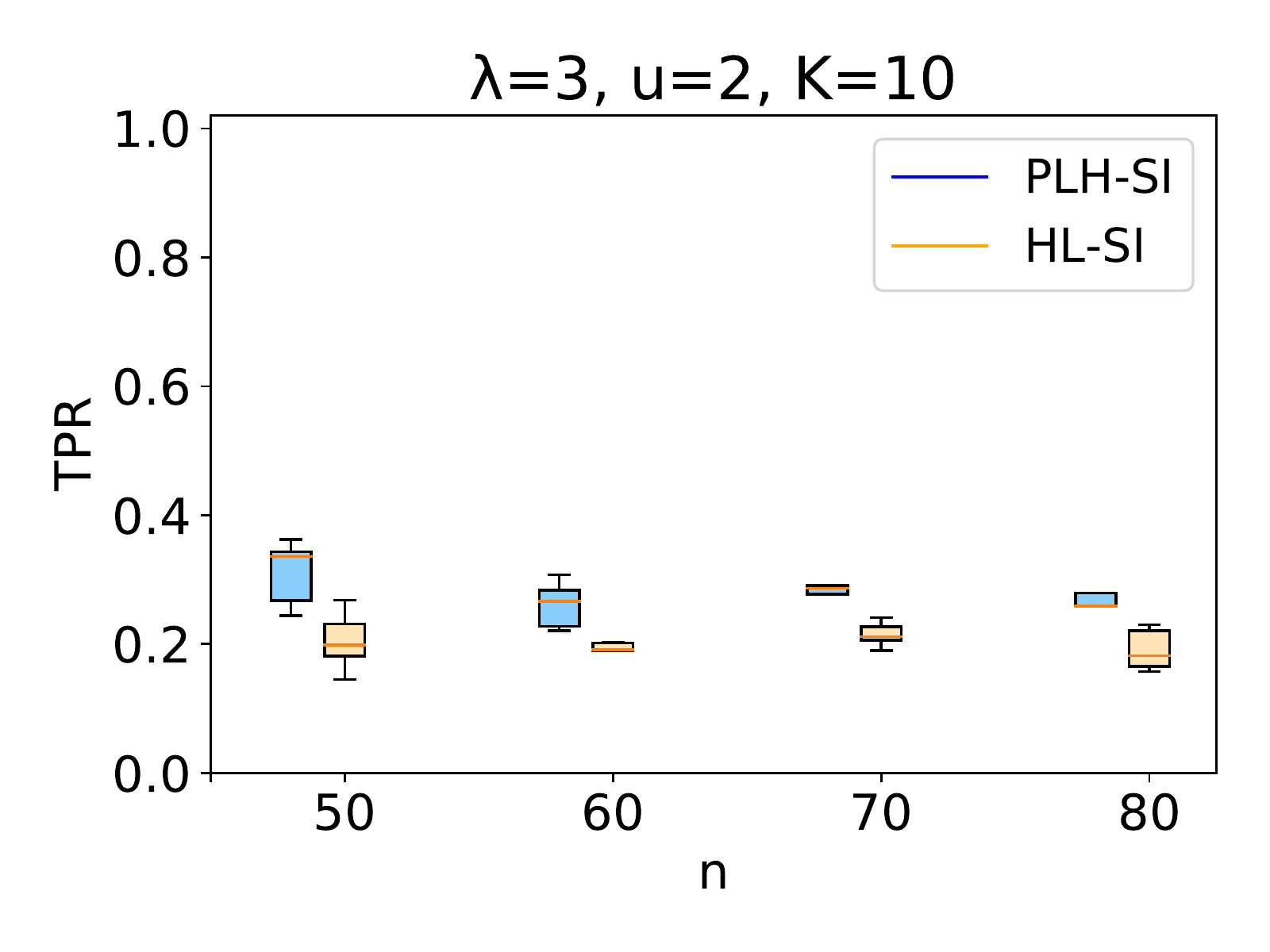}
   &
   \includegraphics[width=0.33\textwidth]{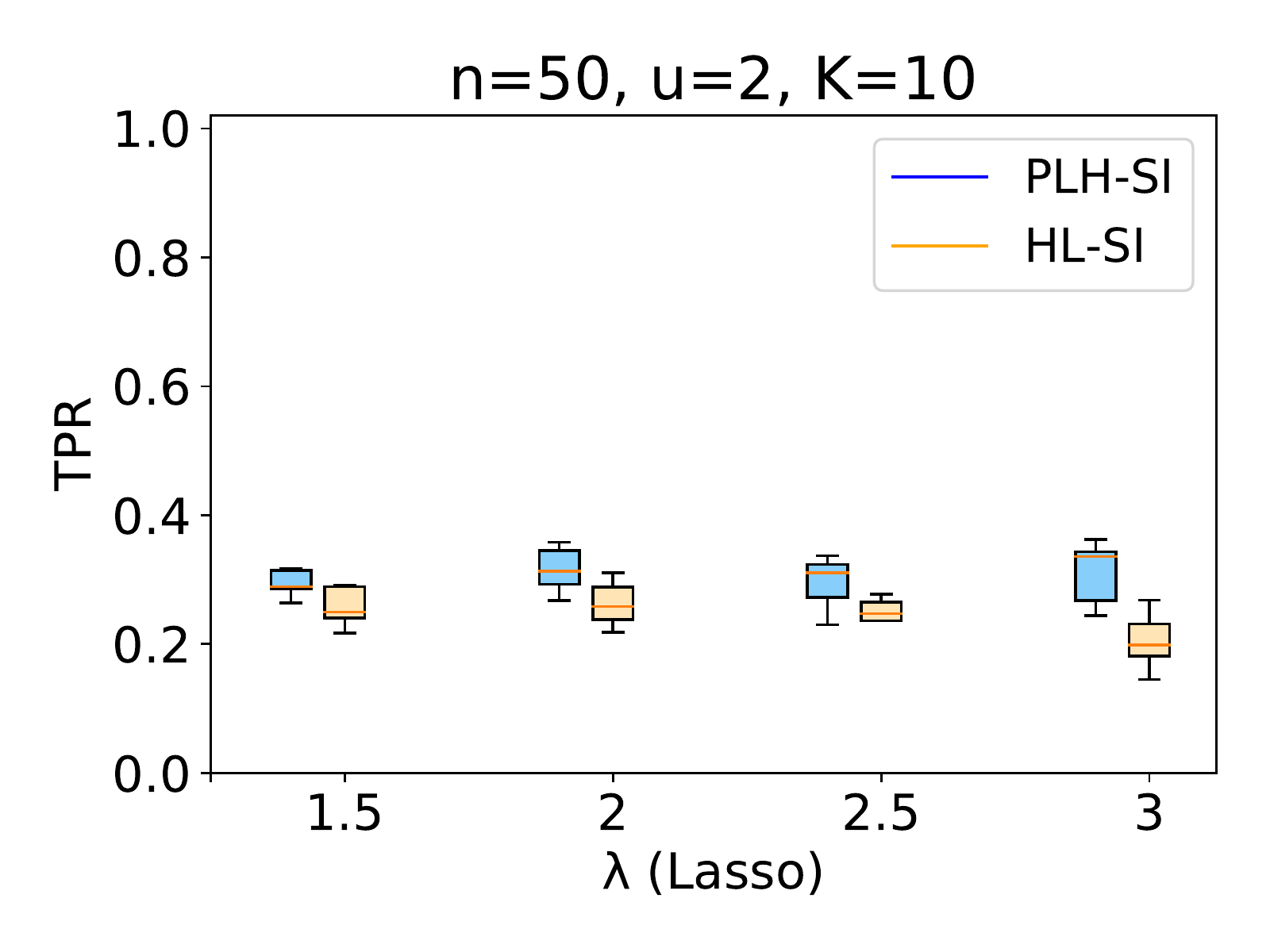}
   \\
   (c) Comparison for different $n$s
   &
   (d) Comparison for different $\lambda$s
  \end{tabular}
  \caption{True positive rates of {\tt PLH-SI} (proposed) and over-conditioned Huberized Lasso-based SI ({\tt HL-SI}) in various setups.}
  \label{fig:homotompy_oc}
 \end{center}
\end{figure}

In this experiment, we demonstrate the difference in terms of TPR between over-conditioned Huberized Lasso, which is discussed in \cite{chen2020valid}, and our proposed method.
We set the first $K$ instances to be outliers by setting the location-shift parameters as $u_i \neq 0$ for all $i \in [K]$, and $u_{i^\prime \in [n] \setminus [K]} = 0$ for the rest of instances, which were not regarded as outliers.
For a fair comparison, the number of detected outliers between Huberized Lasso and the proposed method must be the same.
This requirement is obtained by setting $\delta =\xi= \lambda$, where $\lambda$ is the regularization parameter in Lasso.
The results are shown in Figure~\ref{fig:homotompy_oc}.
We set $n = 50, \lambda = 3, K = 10$, and $u_{i \in [K]} = 2$ as default settings.
In Figure~\ref{fig:homotompy_oc}, we varied $u_{i \in [K]} \in \{2.0, 2.5, 3.0, 3.5\}$ in (a), $K \in \{5, 10, 15, 20\} $ in (b), $n \in \{50, 60, 70, 80\}$ in (c) and $\lambda \in \{1.5, 2.0, 2.5, 3.0\}$ in (d).
Overall, these results are consistent with the discussions in \ref{subsec:discussion_on_related_works}.
The Huberized Lasso has lower TPRs due to over-conditioning on signs.
The details of over-conditioning on signs are presented in Appendix C.

\section{Conclusions}
In this paper, statistical inference of outliers identified by robust regression estimators is considered in the framework of conditional selective inference (SI).
We proposed a practical conditional SI method for this problem by introducing the techniques developed in piecewise-linear homotopy method. 
Numerical experiments show that the proposed method correctly controls the false positive rates and has higher detection power than existing methods.

\clearpage

\appendix

\section{Proofs}
\label{app:proofs}

\paragraph{Proof of Lemma~\ref{lemm:lad_regression}}
Let $\rho_i(z) = y_i(z)-\bm{x}_i^\top\bm{\beta}$ be a parametric residual for each $i$.
Then, (\ref{eq:LAD}) can be written as
\begin{align*}
\hat{\bm{\beta}}^{R}(z) = \argmin_{\bm{\beta}}  \sum_{i=1}^n |\rho_i(z)|.
\end{align*}
Furthermore, for each $i$, let $\rho_i^+(z)$ and $\rho_i^-(z)$  be non-negative numbers with  $\rho_i(z) =\rho_i^+(z)-\rho_i^-(z)$.
Since at least one of $\rho_i(z)^+$ and  $\rho_i(z)^-$ can be set to zero, the following equation holds:
\begin{align*}
|\rho_i(z)|=|\rho_i^+(z) -\rho_i^-(z)| = \rho_i^+(z) +\rho_i^-(z),\ i \in [n].
\end{align*}
Therefore, $\hat{\bm{\beta}}^{R}(z)$ in the parametric approach can be given by  solving the following parametric linear programming problem as
\begin{align}
\label{eq:LP}
&{\rm min} \sum_{i=1}^n\left(\rho_i^+(z) +\rho_i^-(z)\right)\\
&{\rm s.t.}\;\; \rho_i^+(z) - \rho_i^-(z)=y_i(z)-\bm{x}_i^{\top}\bm{\beta}(z), \rho_i^+(z) \ge  0, \rho_i^-(z) \ge  0,\ i \in [n].
\nonumber
\end{align}

Next, we show that (\ref{eq:LP}) can be expressed as (\ref{eq:parametric_linear_program1}).
The $\bm{\beta}(z)$ in (\ref{eq:LP}) can be re-written as follows:
\begin{align*}
\bm{\beta}(z)=\bm{\beta}^+(z) - \bm{\beta}^-(z) , \;\;\; \bm{\beta}^+(z)\ge \bm{0},\;\; \bm{\beta}^-(z)\ge \bm{0}.
\end{align*}
Thus, since ${\bm {Y}} = {\bm a} + {\bm b} z$, by letting 
{\small
\begin{align*}
\bm{r} &= \left(
\begin{array}{cccccccccc}
\rho_1^+ (z) &\rho_1^- (z) & \cdots& \rho_n^+ (z) & \rho_n^- (z)& \beta_{1}^{+} (z) & \beta_{1}^{-} (z) & \cdots& \beta_{p}^{+} (z) & \beta_{p}^{-} (z)
\end{array}
 \right)^\top \in \mathbb{R}^{2n+2p},
\end{align*}
}
%
%
{\footnotesize
\begin{align*}
  S &= \left(
    \begin{array}{cccccccccccccc}
      1 & -1 & 0 &0&\cdots & 0 &0  &x_{11} &-x_{11}&x_{12} &-x_{12}&\cdots &x_{1p}&-x_{1p}\\
      0 & 0 & 1 & -1 &\cdots & 0 &0  &x_{21} &-x_{21}&x_{22} &-x_{22}&\cdots &x_{2p}&-x_{2p}\\
	\multicolumn{2}{c}{\vdots}&\multicolumn{3}{c}{\ddots}& \multicolumn{2}{c}{\vdots}&\multicolumn{2}{c}{\vdots}&\multicolumn{2}{c}{\vdots}& &\multicolumn{2}{c}{\vdots} \\
      0& 0 & 0 &0 &\cdots &1 &-1 &x_{n1} &-x_{n1}&x_{n2} &-x_{n2}&\cdots &x_{np}&-x_{np}\\ 
    \end{array}
  \right) \in \mathbb{R}^{n \times (2n+2p)},\\
\bm{q} &= \left(
\begin{array}{cccccccccc}
1&1&\cdots&1&1&0&0&\cdots&0&0 \\
\end{array}
\right)^\top \in \mathbb{R}^{2n+2p}, 
\bm{u}_0=\bm{a}, \bm{u}_1=\bm{b},
\end{align*}
}
we have  (\ref{eq:parametric_linear_program1}).
Finally, we show piecewise linearity.
It is known that the optimal value of the parametric linear programming problem given in the form (\ref{eq:parametric_linear_program1}) is a (convex) piecewise-linear function with respect to the parameter $z$ (see, e.g., Section 8.6 in \cite{murty1983linear}). 
Hence, the  solution path of $\bm{\beta}^{R}(z)$ can be represented as a piecewise-linear function with respect to $z$.

\paragraph{Proof of Lemma~\ref{lemm:huber_regression}}
%
%
%
%
Let $\bm{e}(z) = \bm{y}(z)-X^\top \bm{\beta}$ be a parametric residual vector.
%
%
Suppose that  
$\bm{u}(z)$ and $\bm{v}(z)$ are vectors, and the $i$-th element $u(z)_i$ and $v(z)_i$ of  $\bm{u}(z)$ and $\bm{v}(z)$ are respectively given by 
$$
u(z)_i = \min \{  |e (z)_i |, \delta\}, \ v(z)_i = \max\{|e (z)_i | -\delta,0 \},
$$
where $e(z)_i$ is the $i$-th element of ${\bm \rho} (z)$. 
Then, the following inequality holds:
\begin{align*}
-\bm{u}(z)-\bm{v}(z) \le \bm{y}(z)-X^\top \bm{\beta} \le \bm{u}(z)+\bm{v}(z).
\end{align*}
In addition, the objective function can be expressed as follows:
\begin{align*}
\frac{1}{2} \bm{u}(z)^\top \bm{u}(z) + \delta \cdot \bm{1}^\top \bm{v}(z),
\end{align*}
where $\bm{1}$ is a vector with all elements $1$.
%
Thus, $\hat{\bm{\beta}}^{R}(z)$ in the parametric approach can be  given by solving the following parametric quadratic programming problem as
\begin{align}
\label{eq:QP}
\min & \frac{1}{2} \bm{u}(z)^\top \bm{u}(z) + \delta \cdot \bm{1}^\top \bm{v}(z)\nonumber\\
\mathrm{s.t.} & -\bm{u}(z)-\bm{v}(z) \le \bm{y}(z)-X\bm{\beta}(z) \le \bm{u}(z) + \bm{v}(z), \\
& \bm{0} \le \bm{u}(z) \le \delta \cdot \bm{1},\;\; \bm{v}(z) \ge {\bm 0}.\nonumber
\end{align}
Therefore, noting that ${\bm y} (z) = {\bm a} + {\bm b} z$,  letting 
\begin{align*}
\bm{r} &= \left(
\begin{array}{ccc}
\bm{\beta} (z) ,\bm{u} (z) ,\bm{v} (z)
\end{array}
\right)^\top \in \mathbb{R}^{p+2n}, \\
  P &= \left(
    \begin{array}{ccc}
      \bm{0}  &\bm{0} &\bm{0}\\
      \bm{0} & I_n & \bm{0}\\
	\bm{0} & \bm{0} & \bm{0}\\
\end{array}
\right)\in \mathbb{R}^{(p+2n) \times (p+2n)}, \\
\bm{q}&=(\bm{0} \;\; \bm{0}\;\;  \delta \cdot \bm{1}_n)^\top \in \mathbb{R}^{(p+2n)}\\
  S &= \left(
    \begin{array}{ccc}
      X  &-I_n &-I_n\\
      -X & -I_n & -I_n\\
	\bm{0} & -I_n & \bm{0}\\
	\bm{0} & I_n & \bm{0}\\
	\bm{0} & \bm{0} & -I_n
\end{array}
\right)\in \mathbb{R}^{5n \times (p+2n)},\\
{\bm u}_0 &=  ({\bm a}  \;\;   -{\bm a} \;\;   \bm{0} \;\; \delta \cdot \bm{1_n} \;\; \bm{0} ) ^\top \in    \mathbb{R}^{5n},  \\
{\bm u}_1 &= ({\bm b}  \;\;   -{\bm b} \;\;   \bm{0} \;\; \bm{0} \;\; \bm{0} )  ^\top \in    \mathbb{R}^{5n},  
\end{align*}
it can be  shown that (\ref{eq:QP}) is the same as  (\ref{eq:parametric_quadratic_program}).

Next, we consider the KKT optimality conditions of the reformulated optimization problem (21), and show that the optimal solutions are represented as piecewise-linear functions of $z$ by showing that, when the active variables (nonzero variables at the optimal solution) do not change, the optimal solutions linearly change with $z$ (Proposition 1). 
Let $\hat{\bm{u}}(z)$ and $\hat{\bm r} (z)$ be the vectors of optimal Lagrange multipliers. 
Then, the KKT conditions of (\ref{eq:parametric_quadratic_program}) are given by
\begin{align*}
P \bm{\hat{r}}(z)+\bm{q}+S^\top \hat{\bm{u}}(z) &={\bm 0},\\
S\hat{\bm{r}}(z)-\bm{h}(z) &\le {\bm 0}, \\
\hat{u}_i (z) (S \hat{\bm{r}}(z)-{\bm h} (z))_i & = 0, \;\;\; \forall i \in [m], \\
\hat{u} _i (z)& \ge 0, \;\;\; \forall i \in [m],
\end{align*}
where $\bm{h}(z)=(\bm{y}(z)\;\;-\bm{y}(z) \;\;  \bm{0} \;\; \delta \cdot \bm{1_n} \;\; \bm{0})^\top \in \mathbb{R}^{5n}$.
Furthermore, letting    $\mathcal{A}_z=\{i \in [m] : \hat{u}_i(z)> 0\}$ and $\mathcal{A}_z^c=[m] \; \backslash \; \mathcal{A}_z$, we have
\begin{align}
\label{kkteq}
P \bm{\hat{r}}(z)+\bm{q}+S^\top \hat{\bm{u}}(z)&={\bm 0}, \nonumber\\
(S\hat{\bm{r}}(z)-\bm{h}(z))_i&= 0, \;\;\; \forall i \in \mathcal{A}_z \\
(S\hat{\bm{r}}(z)-\bm{h}(z))_i&\le 0, \;\;\; \forall i \in \mathcal{A}_z^c. \nonumber
\end{align}
Then, the following proposition holds:
\begin{prop}
\label{lemm:huber}
Let $S_{\mathcal{A}_z}$ be the rows of matrix $S$ in a set $\mathcal{A}_z$. Consider two real values $z$ and $z'(z < z')$.
If $\mathcal{A}_z = \mathcal{A}_{z'}$, then we have
\begin{align}
\label{eqr}
\hat{\bm{r}}(z')-\hat{\bm{r}}(z)&= \psi(z) \times (z'-z), \\
\label{equ}
\hat{\bm{u}}_{\mathcal{A}_z}(z')-\hat{\bm{u}}_{\mathcal{A}_z}(z)&= \gamma(z) \times (z'-z),
\end{align}
where $\psi(z) \in \mathbb{R}^n, \gamma(z) \in \mathcal{R}^{|\mathcal{A}_z|}$, 
$
\begin{bmatrix}
\psi(z)\\
\gamma(z)
\end{bmatrix}
=
\left[
    \begin{array}{cc}
P & S_{\mathcal{A}_z}^\top\\
S_{\mathcal{A}_z} & 0 
    \end{array}
  \right]^{-1}
\left[
\begin{array}{c}
\bm{0}\\
\bm{u}_{1,\mathcal{A}_z}
\end{array}
\right].
$
\end{prop}
Proposition \ref{lemm:huber} means that  $\hat{\bm{\beta}}^{R}(z)$ is a piecewise-linear function with respect to $z$ on the interval $[z,z^\prime]$. 
Finally, we show Proposition  \ref{lemm:huber}. 
\begin{proof}
According to (\ref{kkteq}), we have the following linear system
\begin{align}
\left[
    \begin{array}{cc}
P & S_{\mathcal{A}_z}^\top\\
S_{\mathcal{A}_z} & \bm{0} 
\end{array}
\right]
\left[
    \begin{array}{c}
\hat{\bm{r}}(z)\\
\hat{\bm{u}}_{\mathcal{A}_z}(z)\\
\end{array}
\right]
=
\left[
    \begin{array}{c}
\bm{q}\\
\bm{h}_{\mathcal{A}_z}(z)
\end{array}
\right]. \label{eq:qhA}
\end{align}
By decomposing $\bm{h}(z)=\bm{u}_0+\bm{u}_1 z$,  \eqref{eq:qhA} can be re-written as
\begin{align*}
&\left[
    \begin{array}{cc}
P & S_{\mathcal{A}_z}^\top\\
S_{\mathcal{A}_z} & \bm{0} 
\end{array}
\right]
\left[
    \begin{array}{c}
\hat{\bm{r}}(z)\\
\hat{\bm{u}}_{\mathcal{A}_z}(z)\\
\end{array}
\right]
=
\left[
    \begin{array}{c}
\bm{q}\\
\bm{u}_{0,\mathcal{A}}(z)
\end{array}
\right]
+
\left[
    \begin{array}{c}
\bm{0} \\
\bm{u}_{1,\mathcal{A}_z}
\end{array}
\right] z,
\\
\left[
    \begin{array}{c}
\hat{\bm{r}}(z)\\
\hat{\bm{u}}_{\mathcal{A}_z}(z)\\
\end{array}
\right]
=
&\left[
    \begin{array}{cc}
P & S_{\mathcal{A}_z}^\top\\
S_{\mathcal{A}_z} & \bm{0} 
\end{array}
\right]^{-1}
\left[
    \begin{array}{c}
\bm{q}\\
\bm{u}_{0,\mathcal{A}}(z)
\end{array}
\right]
+
\left[
    \begin{array}{cc}
P & S_{\mathcal{A}_z}^\top\\
S_{\mathcal{A}_z} & \bm{0} 
\end{array}
\right]^{-1}
\left[
    \begin{array}{c}
\bm{0} \\
\bm{u}_{1,\mathcal{A}_z}
\end{array}
\right] z.
\end{align*}
From now, let us denote
$
\begin{bmatrix}
\psi(z)\\
\gamma(z)
\end{bmatrix}
=
\left[
    \begin{array}{cc}
P & S_{\mathcal{A}_z}^\top\\
S_{\mathcal{A}_z} & \bm{0} 
\end{array}
\right]^{-1}
\left[
    \begin{array}{c}
\bm{0} \\
\bm{u}_{1,\mathcal{A}_z}
\end{array}
\right]
$
with $\psi(z)\in \mathbb{R}^n$ and $\gamma(z) \in \mathbb{R}^{|\mathcal{A}_z|}$, the results in proposition \ref{lemm:huber} is straightforward.
\end{proof}

\paragraph{Proof of Lemma~\ref{lemm:threshold_based_outlier_detection}}
We derive the truncation region $\mathcal{Z}$ in the interval $[z_{t-1},z_t].$ 
For any  $t \in [T]$
 and
 $i \in [n]$, the 
residual $r_i^R(z)$ 
in the interval $[z_{t-1},z_t]$ is given by 
\begin{align*}
r_i^R(z)=f_{t,i}+g_{t,i}z \;\; z \in [z_{t-1}, z_t].
\end{align*}
The condition for the truncation region is the following inequality:
\begin{align*}
|r_i^R(z)| & \geq \xi .
\end{align*}
This implies that
\begin{align}
\label{eq:thresholdSE1}
&f_{t,i}+g_{t,i}z \geq \xi, \\
\label{eq:thresholdSE2}
\text{ or } &f_{t,i}+g_{t,i}z \leq -\xi .
\end{align}
From (\ref{eq:thresholdSE1}) and (\ref{eq:thresholdSE2}), $z$  is  classified into the following four cases based on  the value of $g_{t,i}$:
\begin{align*}
&\left\{
    \begin{array}{c}
z\ge \frac{\xi-f_{t,i}}{g_{t,i}} \text{ if } \;g_{t,i} > 0,\\
z\le \frac{\xi-f_{t,i}}{g_{t,i}} \text{ if } \;g_{t,i} < 0,\\
\end{array}
\right. \\
&\left\{
    \begin{array}{c}
z\leq \frac{-\xi-f_{t,i}}{g_{t,i}} \text{ if } \;g_{t,i} > 0,\\
z\geq \frac{-\xi-f_{t,i}}{g_{t,i}} \text{ if }\;g_{t,i} < 0.
\end{array}
\right.
\end{align*}
If $g_{t,i} = 0$, the region that satisfies $|f_{t,i}|\ge \xi$ is the truncation region.
From the above, $\cV_{t, i}$ is derived as
{\small
 \begin{align*}
  \cV_{t, i}
  =
  \mycase{
  \left[
  z_{t-1}, \min\left\{z_{t}, \frac{- \xi - f_{t,i}}{g_{t,i}}\right\}
  \right]
  \cup
  \left[
  \max\left\{z_{t-1}, \frac{\xi - f_{t,i}}{g_{t,i}}\right\}, z_t
  \right]
  &
  \text{ if }
  g_{t,i} > 0,
  \\
  \left[
  z_{t-1}, \min\left\{z_{t}, \frac{\xi - f_{t,i}}{g_{t,i}}\right\}
  \right]
  \cup
  \left[
  \max\left\{z_{t-1}, \frac{-\xi - f_{t,i}}{g_{t,i}}\right\}, z_t
  \right]
  &
  \text{ if }
  g_{t,i} <  0,
  \\
  \left[
  z_{t-1}, z_{t}
  \right]
  &
  \text{ if }
  g_{t,i} = 0
  \text{ and }
  |f_{t,i}| \ge \xi,
  \\
  \emptyset
  &
  \text{ if }
  g_{t,i} = 0
  \text{ and }
  |f_{t,i}| < \xi ,
  }
 \end{align*}
}
where we define 
 $[\ell, u] = \emptyset$ 
 if 
 $\ell > u$.
\paragraph{Proof of Lemma~\ref{lemm:topK_outlier_detection}}
We derive the truncation region $\mathcal{Z}$ in the interval $[z_{t-1},z_t]$.
For any $t \in [T]$
 and
 $(i, i^\prime) \in \cO(y^{\rm observed}) \times \left([n] \setminus \cO(y^{\rm observed})\right)$, 
the residual  
 $r_i^R(z)$ 
in the interval $[z_{t-1},z_t]$ is given by
\begin{align*}
r_i^R(z)=f_{t,i}+g_{t,i}z,\;\;\; r_{i'}^R(z)=f_{t,i'}+g_{t,i'}z \;\; z \in [z_{t-1}, z_t].
\end{align*}
The condition of the truncation region is the following:
\begin{align}
\label{eq:topKSE}
|f_{t,i}+g_{t,i} z| \ge |f_{t,i'} +g_{t,i'}z|.
\end{align}
Since the case classification by absolute value is complicated, we take the method of squaring both sides:
\begin{align}
&(f_{t,i}+g_{t,i} z)^2 \ge (f_{t,i'} +g_{t,i'}z)^2,\nonumber\\
&(g_{t,i}^2-g_{t,i'}^2)z^2 + (2f_{t,i}g_{t,i}-2f_{t,i'}g_{t,i'})z + (f_{t,i'}^2-f_{t,i'}^2) \ge 0,\nonumber\\
\label{eq:Secondary-inequality}
&\alpha z^2 + \beta z + \gamma \ge 0,
\end{align}
where $\alpha=g_{t,i}^2-g_{t,i'}^2, \beta=2f_{t,i}g_{t,i}-2f_{t,i'}g_{t,i'} $ and $\gamma=f_{t,i}^2-f_{t,i'}^2$. 
The truncation region that satisfies \eqref{eq:Secondary-inequality} is equivalent 
to the condition that satisfies equation (\ref{eq:topKSE}). 
Next, we consider the three cases of $\alpha$. 

When $\alpha=0$, 
from $\beta z + \gamma \geq 0$, z is classified into
\begin{align*}
\left\{
    \begin{array}{c}
z\ge \frac{-\gamma}{\beta} \text{ if } \; \beta > 0,\\
z\le \frac{-\gamma}{\beta} \text{ if }\; \beta < 0.
\end{array}
\right. 
\end{align*}
If $\beta=0$, the region that satisfies $\gamma \ge 0$ is the truncation region.
Therefore, when $\alpha = 0$, the truncation region can be expressed as
\begin{align*}
  \mycase{
  \left[
  \max\left\{z_{t-1}, \frac{-\gamma}{\beta}\right\}, z_t
  \right]
  &
  \text{ if }
 \alpha=0 \text{ and } \beta>0,
  \\
  \left[
  z_{t-1}, \min\left\{z_{t},\frac{-\gamma}{\beta}\right\}
  \right]
  &
  \text{ if }
 \alpha=0 \text{ and } \beta<0,
 \\
 \left[z_{t-1}, z_{t}
 \right]
  &
  \text{ if }
 \alpha=0 \text{ and } \beta=0 \text{ and } \gamma\ge0,
\\
\emptyset
  &
  \text{ if }
  \alpha=0 \text{ and } \beta=0 \text{ and } \gamma<0.
 }
 \end{align*}

When $\alpha>0$, 
using the quadratic formula, inequality (\ref{eq:Secondary-inequality}) is
\begin{align*}
&z>\frac{-\beta+\sqrt{\beta^2-4 \alpha \gamma}}{2 \alpha},\\
\text{ or } &z<\frac{-\beta-\sqrt{\beta^2-4 \alpha \gamma}}{2 \alpha}.
\end{align*}
Therefore, the truncation region can be  derived using the union as
{\footnotesize
\begin{align*}
 \left[
  \max\left\{z_{t-1}, \frac{-\beta+\sqrt{\beta^2-4 \alpha \gamma}}{2 \alpha} \right\}, z_t
  \right]
  \cup
  \left[
  z_{t-1}, \min\left\{z_{t}, \frac{-\beta-\sqrt{\beta^2-4 \alpha \gamma}}{2 \alpha}\right\}
  \right]
  \text{ if }
 \alpha>0.
\end{align*}
}

When $\alpha<0$, 
using the quadratic formula, inequality (\ref{eq:Secondary-inequality}) is
\begin{align*}
&z<\frac{-\beta-\sqrt{\beta^2-4 \alpha \gamma}}{2 \alpha}, \\
\text{ and } &z>\frac{-\beta+\sqrt{\beta^2-4 \alpha \gamma}}{2 \alpha}.
\end{align*}
Therefore, the truncation region can be derived using the intersection as 
{\footnotesize
\begin{align*}
 \left[
z_{t-1}, 
  \min\left\{z_{t}, \frac{-\beta-\sqrt{\beta^2-4 \alpha \gamma}}{2 \alpha} \right\}
  \right]
  \cap
  \left[
   \max\left\{z_{t-1}, \frac{-\beta+\sqrt{\beta^2-4 \alpha \gamma}}{2 \alpha}\right\}
, z_{t}
  \right]
  \text{ if }
 \alpha<0.
\end{align*}
}

From the above, $\cW_{t, i}$ is derived as
{\footnotesize
\begin{align*}
  \cW_{t, (i, i^\prime)}
  =
  \mycase{
 \left[
  \max\left\{z_{t-1}, \frac{-\gamma}{\beta}\right\}, z_t
  \right]
  &
  \text{ if }
 \alpha=0 \text{ and } \beta>0,
  \\
  \left[
  z_{t-1}, \min\left\{z_{t},\frac{-\gamma}{\beta}\right\}
  \right]
  &
  \text{ if }
 \alpha=0 \text{ and } \beta<0,
 \\
 \left[z_{t-1}, z_{t}
 \right]
  &
  \text{ if }
 \alpha=0 \text{ and } \beta=0 \text{ and } \gamma\ge0,
\\
\emptyset
  &
  \text{ if }
  \alpha=0 \text{ and } \beta=0 \text{ and } \gamma<0,
\\
 \left[
  \max\left\{z_{t-1}, \frac{-\beta+\sqrt{\beta^2-4 \alpha \gamma}}{2 \alpha} \right\}, z_t
  \right]
  \cup
  \left[
  z_{t-1}, \min\left\{z_{t}, \frac{-\beta-\sqrt{\beta^2-4 \alpha \gamma}}{2 \alpha}\right\}
  \right]
 &
  \text{ if }
 \alpha>0,\\
  \left[
z_{t-1}, 
  \min\left\{z_{t}, \frac{-\beta-\sqrt{\beta^2-4 \alpha \gamma}}{2 \alpha} \right\}
  \right]
  \cap
  \left[
   \max\left\{z_{t-1}, \frac{-\beta+\sqrt{\beta^2-4 \alpha \gamma}}{2 \alpha}\right\}
, z_{t}
  \right]
 &
  \text{ if }
 \alpha<0.
  }
 \end{align*}
}

\section{Breakpoints in Huber regression}
\label{app:breakpoint_calculation}
In this section, we give the lemma for calculating breakpoints. 
Let $\mathcal{A}_z$, $S_{\mathcal{A}_z^c}$,  $\hat{\bm{r}}(z)$, $\hat{\bm{u}}_{\mathcal{A}_z} (z)$,  $\bm{h}_{\mathcal{A}_z^c}(z)$, $\psi (z)$ and  $\gamma (z) $ be the same definitions as in Appendix \ref{app:proofs}.
Then, the following lemma holds:

\begin{lemm}
Consider a real value $z$. Then, 
$\mathcal{A}_z=\mathcal{A}_{z'}$ for any real value $z'$ in the interval $[z, z+t_z]$, where $z+t_z$ is
the value of transition point,
\begin{align*}
t_z=&\min\{t_z^1,t_z^2\},\\
t_z^1=\min_{j \in \mathcal{A}_z^c} \left(-\frac{(S_{\mathcal{A}_z^c}\hat{\bm{r}}(z)-\bm{h}_{\mathcal{A}_z^c}(z))_j}{(S_{\mathcal{A}_z^c} \psi(z))_j} \right)_{++} &
\;\;\; and 
\;\;\;
t_z^2 = \min_{j \in \mathcal{A}_z} \left(-\frac{(\hat{u}_{\mathcal{A}_z} (z))_j}{\gamma_j(z)}\right)_{++}.
\end{align*}
Here, $(a)_{++} = a$ if $a \geq 0$, and otherwise $(a)_{++} = + \infty $. 
\end{lemm}
\begin{proof}
We first show how to derive $t_z^1$. From (\ref{eqr}), we have
\begin{align*}
\hat{\bm{r}}(z')=\hat{\bm{r}}(z)+\psi(z) \times (z'-z).
\end{align*}
Then, we need to guarantee
\begin{eqnarray}
\label{eqzz}
S_{\mathcal{A}_z^c}\hat{\bm{r}}(z)-\bm{h}_{\mathcal{A}_z^c}(z) \le 0, \nonumber \\
S_{\mathcal{A}_z^c}(\hat{\bm{r}}(z')+\psi(z) \times (z'-z)) - \bm{h}_{\mathcal{A}_z^c}(z) \le 0, \nonumber \\ 
S_{\mathcal{A}_z^c}\psi(z) \times (z'-z) \le - (S_{\mathcal{A}_z^c}\hat{\bm{r}}(z)- \bm{h}_{\mathcal{A}_z^c}(z)).
\end{eqnarray}
The right hand side of (\ref{eqzz}) is positive since $S_{\mathcal{A}_z^c}\hat{\bm{r}}(z)- \bm{h}_{\mathcal{A}_z^c}(z) \le 0$.
Therefore,  satisfying equation (\ref{eqzz}) implies that 
\begin{align*}
z'-z \le \min_{j \in \mathcal{A}_z^c} \left(-\frac{(S_{\mathcal{A}_z^c}\hat{\bm{r}}(z)- \bm{h}_{\mathcal{A}_z^c}(z))_j}
{(S_{\mathcal{A}_z^c}\psi(z))_j} \right)_{++} = t_z^1.
\end{align*}
Next, we show how to derive $t_z^2$. From (\ref{equ}), we have
\begin{align*}
\hat{\bm{u}}_{\mathcal{A}_z}(z')=\hat{\bm{u}}_{\mathcal{A}_z}(z) + \gamma(z) \times (z'-z).
\end{align*}
Thus, noting that $\hat{\bm{u}}_{\mathcal{A}_z}(z') \geq 0$ we need guarantee
\begin{eqnarray}
\label{equzz}
\hat{\bm{u}}_{\mathcal{A}_z}(z') = \hat{\bm{u}}_{\mathcal{A}_z}(z) + \gamma(z) \times (z'-z) \geq 0.
\end{eqnarray}
Hence,   satisfying equation (\ref{equzz}) means that
\begin{align*}
z'-z \leq \min_{j \in \mathcal{A}_z} \left(-\frac{(\hat{u}_{\mathcal{A}_z} (z))_j}{\gamma_j(z)}\right)_{++} = t_z^2.
\end{align*}
\end{proof}

\section{Huberized Lasso}
\label{app:huberized_lasso}

According to Equation (8) in \cite{chen2020valid}, we consider the following optimization problem
\begin{align}\label{eq:huberized_lasso_1}
 (\hat{\bm \beta}, \hat{\bm u}) = \argmin \limits_{\bm \beta \in \RR^p, \bm u \in \RR^n} \frac{1}{2} \|\bm y - X \bm \beta - \bm u\|^2_2 + \lambda \|\bm u\|_1,
\end{align}
given $X = (x_1, ..., x_n)^\top$ and $\bm y = (y_1, ..., y_n)^\top$.
Following Sec 3.2 in the supplementary of \cite{chen2020valid}, the optimization in (\ref{eq:huberized_lasso_1}) can be transformed to 
\begin{align*}
 \hat{\bm u} = \argmin \limits_{\bm u \in \RR^n} \frac{1}{2} \|  \tilde{\bm y} -  \tilde{X} \bm u\|^2_2 + \lambda \|\bm u\|_1,
\end{align*}
where $\tilde{\bm y} = P_X^\bot \bm y$ and $\tilde{X} = P_X^\bot$.
We can obtain $\hat{\bm u}$ in (\ref{eq:huberized_lasso_1}) by using Lasso algorithm $\cA$. 
Then, the set of the observed outliers is defined as 
\begin{align*}
 \cA(\bm y) = \{ j : \hat{u}_j \neq 0\}.
\end{align*}
Finally, the inference for a selected outlier is defined as follows 
\begin{align*}
 \bm{\eta}^\top \bm Y \mid \left \{ \cA(\bm Y) = \cA({\bm y}), {\bm q}({\bm Y}) = {\bm q} ({\bm y})\right \}.
\end{align*}
Unfortunately, as pointed out in \cite{lee2016exact}, characterizing $\cA(\bm Y) = \cA({\bm y})$ in (\ref{eq:condition_model}) is computationally intractable because we have to consider $2^{\left |\cA({\bm y}) \right|}$ possible sign vectors.
As suggested in \cite{lee2016exact}, we need to consider inference conditional not only on the selected features but also on their signs to overcome the aforementioned issue.
Specifically, let $\bm s(\bm y)$ denote the sign vector of the selected features when applying Lasso on $\bm y$, the conditional inference we need to focus is 
\begin{equation}\label{eq:condition_model}
	\bm{\eta}^\top \bm Y \mid \left \{ \cA(\bm Y) = \cA({\bm y}), \bm s(\bm Y) = \bm s(\bm y), {\bm q}({\bm Y}) = {\bm q} ({\bm y})\right \}.
\end{equation}
However, additionally considering the signs leads to low statistical power because of \emph{over-conditioning}.
This is well-known as the major drawback in SI literature.

\clearpage

This work was partially supported by MEXT KAKENHI (20H00601, 16H06538), JST CREST (JPMJCR21D3), JST Moonshot R\&D (JPMJMS2033-05), JST AIP Acceleration Research (JPMJCR21U2), NEDO (JPNP18002, JPNP20006) and RIKEN Center for Advanced Intelligence Project. We thank the two anonymous reviewers for their constructive comments which help us to improve the paper. 


\end{document}